\documentclass[conference]{IEEEtran}

\IEEEoverridecommandlockouts

\usepackage{cite}
\usepackage{amsmath,amssymb,amsfonts}

\usepackage{graphicx}
\usepackage{textcomp}
\usepackage[table,xcdraw]{xcolor}
\usepackage{multirow}
\usepackage{booktabs}
\usepackage{soul}
\usepackage{comment}

\usepackage{adjustbox}
\usepackage{graphicx}
\usepackage{float}
\usepackage{pgfplots}
\usetikzlibrary{patterns}
\usepackage[margin=1in]{geometry}
\usepackage{caption}
\usepackage{threeparttable}

\usepackage{xcolor}
\usepackage{balance}

\usepackage{algorithm}
\usepackage{algpseudocode}
\usepackage[most]{tcolorbox}
\usepackage{hhline}

\tcbset{mychangebar/.style={
  enhanced,
  breakable,                  
  frame hidden,
  borderline west={2pt}{0pt}{black}, 
  sharp corners,
  boxrule=0pt,
  colback=white,
  left=6pt, right=0pt, top=0pt, bottom=0pt,
}}

\newenvironment{cbblock}{\begin{tcolorbox}[mychangebar]}{\end{tcolorbox}}
\newcommand{\cbstart}{\begin{cbblock}}
\newcommand{\cbend}{\end{cbblock}}

\def\BibTeX{{\rm B\kern-.05em{\sc i\kern-.025em b}\kern-.08em
    T\kern-.1667em\lower.7ex\hbox{E}\kern-.125emX}}

\title{Autonomous Collaborative Learning Among an Ensemble of Tsetlin Machines with Consensus-Based Inference}

\author{
       Yehuda Rudin, Osnat Keren, Michal Yemini, and Alexander Fish 
       \thanks{Y.~Rudin, M.~Yemini, and A.~Fish are with Faculty of Engineering, Bar Ilan University, Ramat Gan 5290002, Israel. Emails: yehuda.rudin@biu.ac.il, osnat.keren@biu.ac.il, michal.yemini@biu.ac.il, alexander.fish@biu.ac.il.}
    }

\pgfplotsset{compat=1.18}
\begin{document}

\maketitle

\begin{abstract}

Tsetlin Machine (TM) is a rule-based machine-learning algorithm comprising collectives of two-action Tsetlin Automata (TAs) that cooperatively form conjunctive logical clauses from Boolean inputs through stochastic feedback. 
Although few recent studies have examined TM Federated Learning, the broader area of distributed and decentralized TM learning has not received much attention in the existing literature and warrants further exploration.  In this work, we propose a paradigm for decentralized collaborative learning under a vertical feature-partitioning setting among an ensemble of Tsetlin Machines using consensus-based inference. Within this decentralized paradigm, each agent maintains its own private TM model, and there is no exchange of raw data among agents. Inference combines individual agents' model predictions into a global consensus. The paradigm accommodates heterogeneous TM-based agents with differing data acquisition means, local data distributions, or computational resources, thereby facilitating the integration and fusion of information in settings such as multi-modal sensing environments. Experiments conducted using two-dimensional grid and connected graph network topologies demonstrate that the classification accuracies achieved are comparable to those of centralized models.

\end{abstract}


\section{Introduction}
\label{sec: Introduction}
The Tsetlin Machine (TM), first introduced in 2018 \cite{Granmo2018TheTM}, represents a logic-driven machine learning paradigm that serves as an alternative to Artificial Neural Networks (ANNs). Since its introduction, an extensive family of TM variants has emerged, including  \cite{granmo2019ctm, abeyrathna2019regression,jiao2020mctm, abeyrathna2020weighted, abeyrathna2021iwtm, abeyrathna2021cmtm, yadav2021continuous, abeyrathna2021recurrent, yadav2022fuzzy, yazidi2022reltm, tarasyuk2024mltm, Kuruge_2024, 10.3389/frai.2025.1377944, granmo2026tsetlinmachinegoesdeep}, each extending the capabilities of the core framework.

The theoretical foundation of TM can be traced back to the work of Michael Tsetlin, who pioneered the concept of the learning automaton and formulated the theory of \textit{automata collective behavior} \cite{1974automation, PuppetsWithoutStrings, M.L.Tsetlin_1963}. According to the theory, intricate, system-level behavior can emerge from local interactions of agents following simple rules and performing a set of simple actions. Through local interactions alone, without any centralized controller, the group can pursue and attain complex common goals that exceed the capabilities of any individual agent \cite{swarm_intelligence}. This emergent perspective underpins the design philosophy of the Tsetlin Machine.

Concretely, TM builds on the Tsetlin Automaton (TA) and propositional logic; it uses collections of logical clauses constructed from Boolean input features to encode and recognize patterns in data. These clauses, shaped through reinforcement-driven updates of the underlying automata, form interpretable logical expressions that capture relevant input–output relationships.

Because its operations are grounded in discrete, logic-based computations rather than continuous-valued matrix operations, TM is inherently simple from a computational standpoint and can be implemented efficiently in both software and hardware. This simplicity translates into reduced energy consumption \cite{Lei2020FromAT}. Consequently, TM is particularly attractive for resource-constrained and energy-limited environments, such as embedded systems and edge computing platforms \cite{Wheeldon2020LearningAB, 10.1145/3560905.3568512}. Grounded in the framework of propositional logic, the TM's underlying decision-making process can be rationalized and explained by human-readable conjunctive clauses.

Data-generating edge devices typically operate under strict constraints on both computation and energy. In these settings, it is also important to safeguard the privacy of locally collected data. Most machine learning methods rely on complex arithmetic operations and therefore consume substantial energy, posing barriers for edge devices, particularly during model training. One possible approach to circumvent this limitation is to delegate the computationally intensive tasks to cloud-based resources. However, this strategy introduces additional drawbacks, notably increased latency, higher energy expenditure associated with data transmission and network communication, and a compromise of privacy by uploading raw data to the cloud. 
 
Instead, it is desirable to perform learning and inference in proximity to the data sources, exploiting and coordinating the local computational resources available at all participating agents.

Federated Learning (FL) \cite{AbhishekV2022FederatedLC, mcmahan2023communicationefficientlearningdeepnetworks, akhtarshenas2024federatedlearningcuttingedgesurvey} is used when the data is physically distributed, e.g., generated from a network of sensors spatially spread, and when there is a need to preserve data privacy. FL offers a paradigm in which a central server orchestrates model training across a collection of devices. FL is typically classified into Horizontal Federated Learning (HFL) and Vertical Federated Learning (VFL) based on the data distribution between the participating agents \cite{yang2019federatedmachinelearningconcept}.

In HFL, agents share the same feature set but have different samples, whereas in VFL, agents possess different feature sets but the same sample set.  In HFL, a shared model is trained across several agents. A central server coordinates the training by sending an initial model to the clients.  Each client trains the model locally on its private data. The model updates (weight deltas, gradients) are sent back to the server, which aggregates them to improve the global model.  The cycle repeats until the model reaches the desired performance. In the standard workflow, each agent uses the globally trained model independently for inference.

In VFL settings, different agents typically hold different feature subsets for the same samples. Each agent has its own local sub‑model that consumes its own features only and outputs an embedding or partial prediction. A central aggregation model (on a server or central agent) combines these embeddings to produce the final prediction. 
Knowledge Distillation-based approaches \cite{Li_2024} have emerged as the most prevalent in this category, enabling knowledge transfer between models with differing structures. 
Whereas conventional VFL paradigms restrict training to shared samples across participating parties, the approach proposed in \cite{Huang_2023} generalizes this framework by enabling each party to additionally exploit its own private, non-overlapping local samples via a dedicated knowledge transfer mechanism.

VFL is designed for cross-silo environments where multiple parties possess data on the same set of users or entities but operate in distinct domains, are willing to cooperate, and do not have conflicting interests. It is particularly advantageous when data cannot be consolidated into a single repository because of privacy constraints, regulatory requirements, or organizational barriers. Prominent VFL applications include collaborative healthcare analytics, banking, credit risk assessment, and e-commerce \cite{wu2025verticalfederatedlearningpractice}. It is also employed to jointly train large-scale machine learning models in a distributed fashion \cite{JMLR:v22:20-815}.

Decentralized Federated Learning (DFL) \cite{kairouz2021advancesopenproblemsfederated} eliminates the central server; agents only communicate with their neighbors to reach a consensus model. DFL offers a communication advantage and further preserves data privacy. 

FedMD \cite{li2019fedmdheterogenousfederatedlearning} is a heterogeneous federated learning method based on knowledge distillation. Every agent maintains its own private dataset and an individually tailored model. The information learned by each agent is converted into a common representation, which is then aggregated by a central server to derive a consensus that is shared back with the agents.

Federated learning was proposed for the Tsetlin Machine in \cite{Qi2023FedTMMA, qi2025fedtmos, gohari2024tpfltsetlinpersonalizedfederatedlearning}, all of which use a central server for model aggregation.
In this work, we investigate a TM-based decentralized learning framework under a vertical federated learning data partitioning scheme, wherein agents hold disjoint feature subsets of the same data instances and rely on peer-to-peer communication to achieve consensus-based prediction (inference). The agents may be diverse, such that each agent may have its own machine learning model, which is not shared with the other agents, thereby further increasing the heterogeneity of the learning process.

\begin{figure*}[!h]
        \centering
         \includegraphics[width=14cm, trim = {0cm 4.6cm 10.2cm 0cm}, clip]{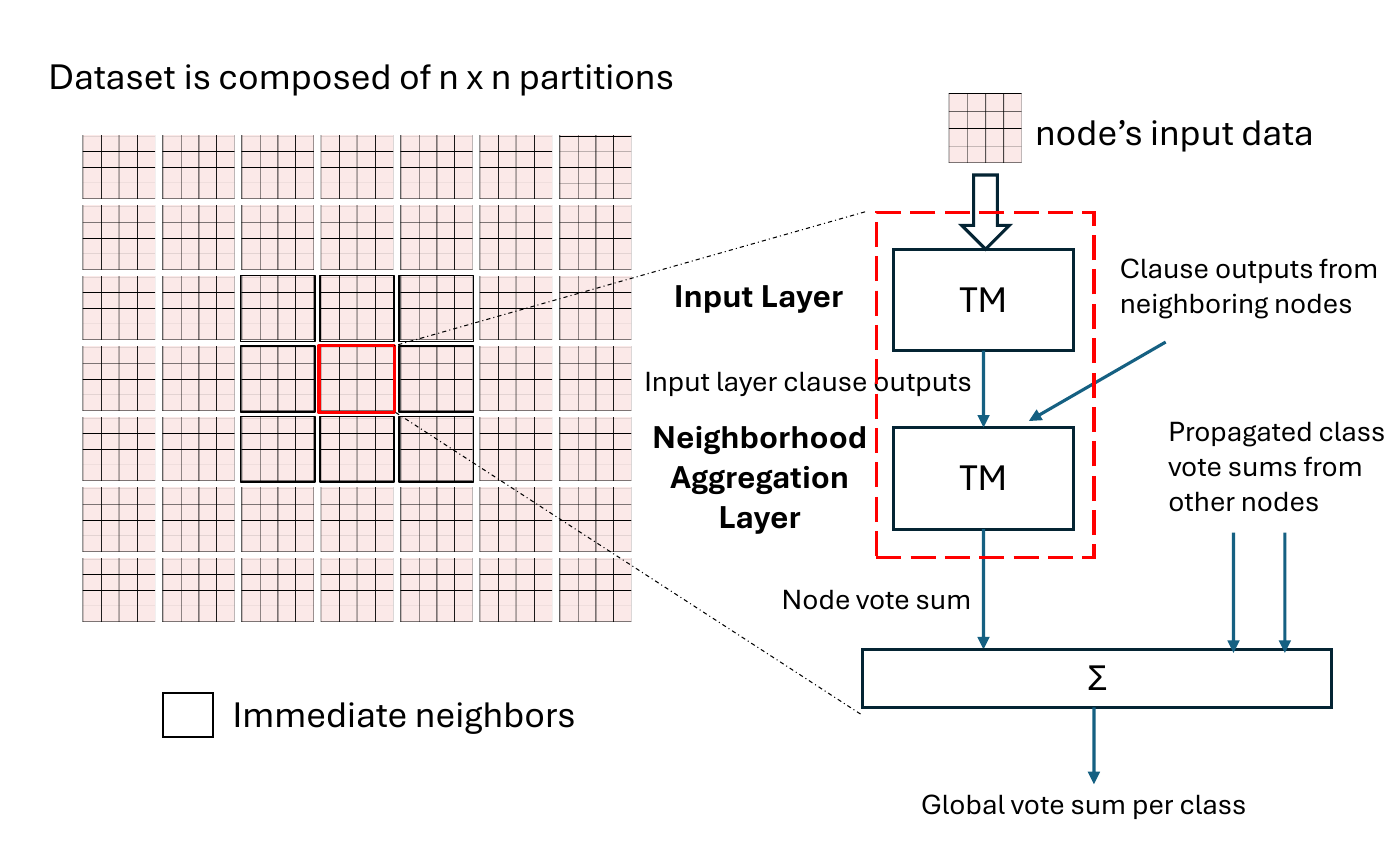}
        \caption{Architecture of an Agent.}
        \label{fig:architecture}
\end{figure*}

Our novel approach is grounded in a two-layer hierarchical TM architecture, to the best of our knowledge, the first of its kind, which is paired with a gossip-based communication protocol. This configuration enables agents to exchange information efficiently with their neighboring peers, while simultaneously precluding the direct transmission of raw data and explicit model parameters, to safeguard the confidentiality of both (formal leakage analysis is left for future work).
At the first layer of the hierarchy, which we name the ``input layer", the TM operates on locally available features and performs classification tasks using only this subset of features. At the second layer, which we name ``the neighborhood aggregation layer", the framework consolidates the rule-based outputs produced by each agent with those obtained from its neighboring agents. Through this aggregation over a cluster of interconnected nodes and an expanded, more heterogeneous feature space, the second tier can recognize and characterize more intricate input patterns that cannot be captured by isolated local models alone. The proposed architecture is presented in Fig. \ref{fig:architecture}.

 \subsection*{Main contributions}
 
 The main contributions of this work are as follows:
 \begin{itemize}
    \item We introduce a hierarchical TM framework that enables distributed, decentralized learning under aligned samples and a vertical feature-partitioning setting with consensus-driven inference. This framework is designed to preserve the confidentiality of both the underlying model parameters and the local data of each participant.
     \item The proposed paradigm accommodates heterogeneous TM-based agents with differing data acquisition means, local data distributions, or computational resources, thereby facilitating the integration and fusion of information in settings such as large-scale sensor networks and multi-modal sensing environments.
     \item We propose an inter-agent asynchronous gossip-based communication protocol for coordination and data exchange.
    \item We analyze the model's performance, using software implementation,  on the MNIST and Fashion-MNIST datasets, partitioned among several agents. 
    \item We analyze the model's performance on a synthetic heterogeneous sensor network setup, in comparison to a centralized ANN model.
    \item We assess the sensor network model under two-dimensional grid and general connected graph network topologies.
     \item We evaluate performance as a function of system configuration variants such as the number of agents (dataset partitions) and the TMs' hyperparameters in each hierarchy.
 \end{itemize}

\subsection*{Organization} The remainder of this paper is structured as follows.
 Section \ref{sec:background} provides an introduction to TM and reviews previous work related to FL using TM, Section \ref{sec: Algorithm Design} describes the proposed algorithm for collaborative decentralized TM learning, and Section \ref{sec:experimentation} presents performance results over the MNIST handwritten digit recognition benchmark, the Fashion-MNIST benchmark, and a heterogeneous sensor network model. Section \ref{sec:Summary} summarizes this work.

\section {Technical Background}
\label{sec:background}

This section outlines the architecture of the TM model, explains the core principles underlying its operation, and specifies its hyperparameter definitions. In addition, it presents techniques used in distributed machine learning settings. 

\subsection{TM Overview}
\label{subsec:TM overview}
\begin{figure*}[ht]
        \centering
        \includegraphics[trim ={0cm, 6.75cm 0cm 0cm}, clip, scale=0.5]{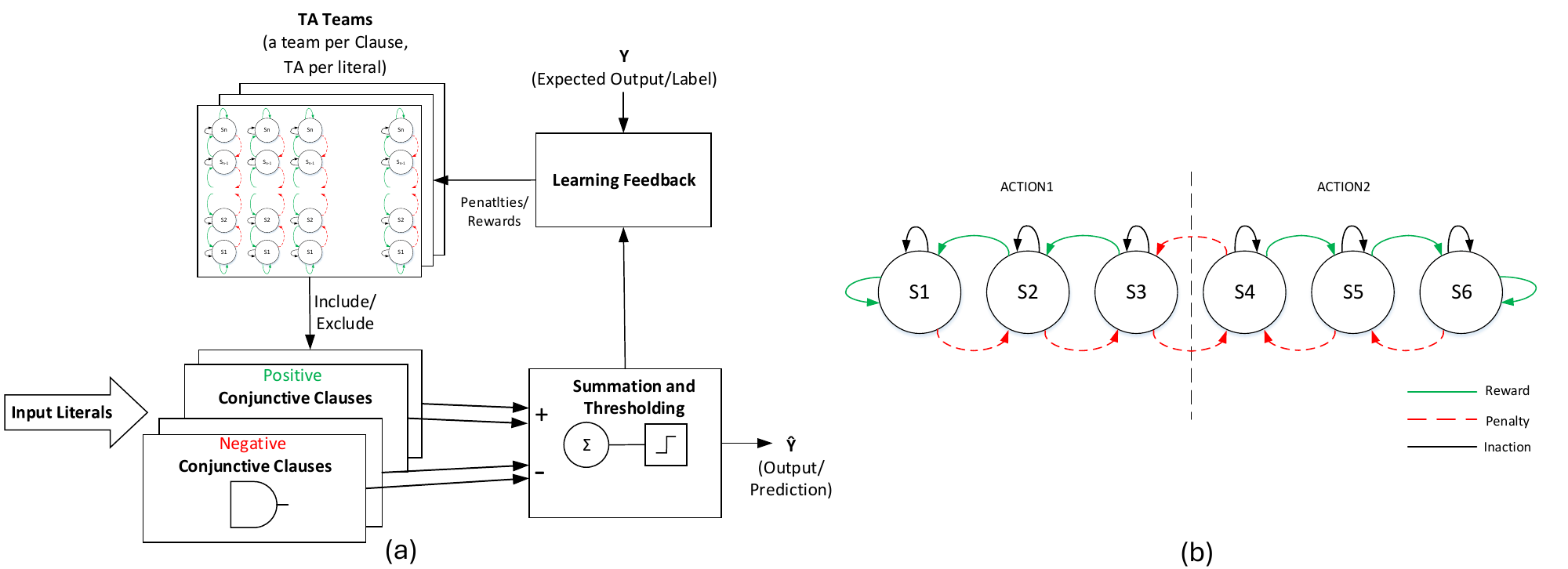}
        \caption{The Tsetlin Machine. (a) Block Diagram of a Tsetlin Machine.  (b) State diagram of Tsetlin learning Automaton (6-state example)}
        \label{fig:TM_TA}
    \end{figure*}

A block diagram of TM is presented in Fig. \ref{fig:TM_TA}(a).

The Tsetlin Machine (TM) performs classification of input features by employing conjunctive clauses to detect discriminative patterns. The input space is represented by Boolean features together with their complements, collectively referred to as literals. Half of the clauses (positive clauses) are dedicated to recognizing patterns associated with the target class, whereas the remaining clauses (negative clauses) are responsible for identifying patterns indicative of non-membership in the target class. 

The final classification decision is obtained by summing the outputs of the positive and negative clauses, taking into account their respective signs, and comparing the resulting aggregate score to a predefined threshold. This procedure effectively implements a voting mechanism over the set of clauses.


Each clause is associated with a team of Tsetlin Automata (TAs), with one automaton assigned to each literal. A TA is a finite-state machine (FSM) employed to construct logical clauses through the inclusion or exclusion of literals. Depending on its current state, a TA chooses between two possible actions: Include or Exclude. For every action executed, the Learning Feedback module produces either a Reward, Penalty, or Inaction, following a probability distribution that depends on the specific literal assigned to the automaton as well as the relationship between the clause output and the expected output. The TA updates its internal state in response to this feedback to rapidly converge to the action associated with the highest reward probability. Figure~\ref{fig:TM_TA}(b) depicts an example of a TA comprising six states.


A Tsetlin Machine is characterized by two principal hyperparameters, denoted by \(T\) and \(s\). The threshold parameter \(T\) regulates the target magnitude of the aggregate clause vote. Consequently, a higher threshold value entails that a greater number of clauses contribute to the voting process and thereby affect the feedback dynamics governing the TA states. During the training phase, the probability of issuing feedback increases as the deviation of the clause vote from the threshold \(T\) grows. In contrast, when the clause vote reaches or surpasses the threshold \(T\), the feedback mechanism is effectively suppressed and no further feedback is provided.


The hyperparameter $s$ regulates the probability with which TAs transition between internal states. It controls the sensitivity (Specificity) of the learning process.  A high value of $s$ promotes the learning of specific patterns by forming specialized clauses, while a low value of $s$ generates more general clauses.

Careful tuning of these parameters can be used to determine learning stability and robustness \cite{Granmo2018TheTM}. 

As illustrated in Fig. \ref{fig:TM_TA}(b), a TA updates its state in response to one of three possible forms of feedback: Reward, Penalty, or Inaction. Two distinct feedback mechanisms are defined within the TM learning framework: Type I feedback and Type II feedback, which are specified in Table \ref{table: type I feedback} and Table \ref{table: type II feedback}, respectively.

\begin{table}[h]
\setlength\extrarowheight{5pt}
\centering
\caption{Type I Feedback}
\label{table: type I feedback}
\begin{tabular}{|l|l|cc|cc|}
\hline 
\multicolumn{1}{|c|}{} &
  \multicolumn{1}{c|}{\textbf{Clause}} &
  \multicolumn{2}{c|}{\textbf{1}} &
  \multicolumn{2}{c|}{\textbf{0}} \\ \cline{2-6} 
\multicolumn{1}{|c|}{\multirow{-2}{*}{\textbf{Action}}} &
  \multicolumn{1}{c|}{\textbf{Literal}} &
  \multicolumn{1}{c|}{\textbf{1}} &
  \textbf{0} &
  \multicolumn{1}{c|}{\textbf{1}} &
  \textbf{0} \\ \hline
                                  & P(reward)   & \multicolumn{1}{c|}{$\frac{s-1}{s}$} & -               & \multicolumn{1}{c|}{0}               & 0               \\ \cline{2-6} 
                                  & P(inaction) & \multicolumn{1}{c|}{$\frac{1}{s}$}   & -               & \multicolumn{1}{c|}{$\frac{s-1}{s}$} & $\frac{s-1}{s}$ \\ \cline{2-6} 
\multirow{-3}{*}{Include Literal} & P(penalty)  & \multicolumn{1}{c|}{0}               & -               & \multicolumn{1}{c|}{$\frac{1}{s}$}   & $\frac{1}{s}$   \\ \hline
                                  & P(reward)   & \multicolumn{1}{c|}{0}               & $\frac{1}{s}$   & \multicolumn{1}{c|}{$\frac{1}{s}$}   & $\frac{1}{s}$   \\ \cline{2-6} 
                                  & P(inaction) & \multicolumn{1}{c|}{$\frac{1}{s}$}   & $\frac{s-1}{s}$ & \multicolumn{1}{c|}{$\frac{s-1}{s}$} & $\frac{s-1}{s}$ \\ \cline{2-6} 
\multirow{-3}{*}{Exclude Literal} & P(penalty)  & \multicolumn{1}{c|}{$\frac{s-1}{s}$} & 0               & \multicolumn{1}{c|}{0}               & 0               \\ \hline
\end{tabular}
\end{table}

Type I feedback reinforces true positive output (Type Ia feedback) and reduces false negative output (Type Ib feedback).  When the clause outputs "1" for the target class, if a TA action is to include a literal valued '1', it receives reward feedback with probability $\frac{s-1}{s}$, otherwise ``inaction" occurs with a probability of $\frac{1}{s}$. If the TA action, in this case, is to exclude a literal valued '0', it receives a reward feedback with probability $\frac{1}{s}$, otherwise, ``inaction" occurs with a probability of $\frac{s-1}{s}$.
In case the clause outputs '0' for the target class, the feedback is directed to combat the false negative output. Given the need for clause reconstruction, the feedback penalizes all inclusions while rewarding all exclusions.

\begin{table}[h]
\setlength\extrarowheight{5pt}
\centering
\caption{Type II Feedback}
\label{table: type II feedback}
\begin{tabular}{|l|l|cc|cc|}
\hline 
\multicolumn{1}{|c|}{} &
  \multicolumn{1}{c|}{\textbf{Clause}} &
  \multicolumn{2}{c|}{\textbf{1}} &
  \multicolumn{2}{c|}{\textbf{0}} \\ \cline{2-6}  
\multicolumn{1}{|c|}{\multirow{-2}{*}{\textbf{Action}}} &
  \multicolumn{1}{c|}{\textbf{Literal}} &
  \multicolumn{1}{c|}{\textbf{1}} &
  \textbf{0} &
  \multicolumn{1}{c|}{\textbf{1}} &
  \textbf{0} \\ \hline
                                  & P(reward)   & \multicolumn{1}{c|}{0} & - & \multicolumn{1}{c|}{0} & 0 \\ \cline{2-6} 
                                  & P(inaction) & \multicolumn{1}{c|}{1} & - & \multicolumn{1}{c|}{1} & 1 \\ \cline{2-6} 
\multirow{-3}{*}{Include Literal} & P(penalty)  & \multicolumn{1}{c|}{0} & - & \multicolumn{1}{c|}{0} & 0 \\ \hline
                                  & P(reward)   & \multicolumn{1}{c|}{0} & 0 & \multicolumn{1}{c|}{0} & 0 \\ \cline{2-6} 
                                  & P(inaction) & \multicolumn{1}{c|}{1} & 0 & \multicolumn{1}{c|}{1} & 1 \\ \cline{2-6} 
\multirow{-3}{*}{Exclude Literal} & P(penalty)  & \multicolumn{1}{c|}{0} & 1 & \multicolumn{1}{c|}{0} & 0 \\ \hline
\end{tabular}
\end{table}

Type II feedback is activated when the clause outputs '1' for a pattern that does not belong to the target class. In this case, the exclusion of a literal that is '0' is penalized.

TM has been shown to be effective in image classification \cite{granmo2019ctm}, IoT applications  \cite{10.1145/3560905.3568512, Wheeldon2020LearningAB, 10455063},  low-power speech recognition \cite{lei2021lowpower}, and Natural Language Processing (NLP) \cite{https://doi.org/10.1111/exsy.12873,bhattarai2023tsetlin,bhattarai2020measuring,Yadav_Jiao_Granmo_Goodwin_2021}.

For reference, Table \ref{tab:ML algorithms}, reproduced from \cite{granmo2019convolutionaltsetlinmachine}, shows the MNIST and Fashion-MNIST test accuracy results, comparing classic TM with various machine learning algorithms. A comparative analysis of TM to Neural Networks regarding learning convergence and energy efficiency is presented in \cite{Lei2020FromAT}. 

\begin{table}[H]
\centering
\caption{Test accuracy in percent for TM and selected popular machine learning algorithms, adapted from \cite{granmo2019convolutionaltsetlinmachine}.}
\label{tab:ML algorithms}
\begin{tabular}{| l | c | c |}
\hline
\textbf{Model} & \multicolumn{1}{l|}{\textbf{MNIST}} & \multicolumn{1}{l|}{\textbf{F-MNIST}} \\ \hhline{|===|}
4-Nearest Neighbors            & 97.14          & 85.40          \\ \hline
SVM                            & 98 .57         & 89 .7          \\ \hline
Random Forest                  & 97 .3          & 81 .6          \\ \hline
Gradient Boosting Classifier   & 96 .9          & 88 .0          \\ \hline
Simple CNN                     & 99 .06         & 90 .7          \\ \hline
Binary Connect                 & 98 .99         & --             \\ \hline
FPGA accelerated BNN           & 98 .70         & --             \\ \hline
Logistic Circuit (binary)      & 97 .4          & 87 .6          \\ \hline
Logistic Circuit (real-valued) & 99 .4          & 91 .3          \\ \hline
PreActResNet-18                & 99 .56         & 92 .00         \\ \hline
ResNet18 + VGG Ensemble        & 99 .60         & --             \\ \hline
\textbf{TM}                    & \textbf{98.57} & \textbf{90.09} \\ \hline
\end{tabular}
\end{table}

\subsection{Federated Learning with Tsetlin Machine}
\label{subsec:distributed ML}
Inspired by conventional Federated Learning with Convolutional Neural Networks (CNNs), FedTM \cite{Qi2023FedTMMA} is a framework that uses Tsetlin Machine in FL.  As in HFL, all FedTM clients use a global model.  The server performs aggregation in two steps: aggregation of clause weights (AverageCW) and aggregation of clause states (TopK). AverageCW calculates the weighted average of integer clause weights for each class across the participating $J$ clients, based on their dataset sizes. 
TopK aggregates the states from a subset of $K$ clients for a specific class. The states are selected from clients with the highest number of samples per class, indicating greater confidence, and are combined using the bitwise OR operator. Compared to conventional Federated Averaging (FedAvg) with CNNs, on average, FedTM provides a substantial reduction in communication costs by 30.5× and 36.6× reduction in memory footprint.

Tsetlin-Personalized Federated Learning with Confidence-Based Clustering (TPFL) \cite{gohari2024tpfltsetlinpersonalizedfederatedlearning} was introduced to tackle a key challenge in federated learning (FL): the heterogeneity of data distributions across clients. In this approach, a client’s model parameters are aggregated only when the client is sufficiently confident in what it has learned, meaning it has seen enough training samples for a given class. This notion of confidence is derived from the voting mechanism of the Tsetlin Machine (TM) algorithm: a larger number of votes in favor of a class indicates higher confidence in the corresponding prediction. 
Each client transmits to the server its weight vector along with the label of the class for which it has the highest confidence. This substantially lowers the communication overhead. The server then partitions the clients into $K$ clusters, where cluster $k$ consists of those clients whose maximum confidence score is associated with class $k$. Within each cluster $k$, the clients’ weight vectors are aggregated.

One-Shot Federated Learning With Tsetlin Machine, FedTMOS \cite{qi2025fedtmos}, restricts communication with the server to a single round, thus minimizing communication errors and reducing the risk of interference caused by iterative updates. Given $J$ clients, each having local datasets $D_1,D_2, ...,D_J$ . The objective is to aggregate local TM models, $T = \{T_1, T_2, ..., T_J\}$, into $\phi$ server models $(\phi < J)$ that generalize well over all datasets in one communication round. The aggregation of the models is done by applying the principles behind TM Composites \cite{granmo2023tmcompositesplugandplaycollaborationspecialized}. In the initial step, clients upload their scaled clause weights, which are adjusted based on the proportion of samples per class relative to the client’s total sample size. Alongside these weights, clients also upload their individual normalized Gini index, which quantifies the inequality in their local data distributions. The server then rescales the weights using the mean normalized Gini Index and performs k-means clustering on the weights.  This inter-class weight separation technique is applied to create models that enhance class distinction. Finally, the $\phi$ server-side models are initialized, and class weights from each cluster are reassigned to maximize inter-class separation within each model.

\subsection{Motivation}
\label{subsec:motivation}
The TM has been introduced as an alternative learning architecture grounded in logic-based, bit-level operations that naturally map to hardware implementations. Due to its reliance on simple, low-complexity computations rather than continuous-valued arithmetic, the TM is particularly well suited to severely resource-constrained and energy-limited environments, including dense sensor networks and other edge-computing scenarios.  Several works on TM-based federated learning have recently been published \cite{Qi2023FedTMMA, qi2025fedtmos, gohari2024tpfltsetlinpersonalizedfederatedlearning}, however, the domain of distributed and decentralized TM learning has, to a large extent, remained unexplored and has received minimal attention in the current literature.
In this work, we consider a model-heterogeneous decentralized learning setting, where agents may employ different local model architectures and collaborate without exposing the local data or the underlying model. We assume that training samples are vertically aligned and that the class labels are available to all agents.

\section{Algorithm Design}
\label{sec: Algorithm Design} 


The proposed distributed and decentralized learning algorithm is defined over an underlying undirected, fixed-in-time, and connected communication graph. The experimental evaluation in this study primarily considers a two-dimensional cellular lattice of size \(n \times n\), in which each cell contains a single agent and is connected to its immediate eight neighboring cells. To eliminate boundary artifacts arising from agents having fewer than eight immediate neighbors, we consider a toroidal wrap-around grid. In addition, we assess the algorithm’s performance on a more general graph network topology.

Each agent (for instance, a sensor node in a distributed sensor network) acquires and processes its own local dataset, which remains strictly private and is never directly shared. During both the training (learning) phase and the prediction (inference) phase, agents cooperate only with their immediate neighbors in the grid topology, exchanging information in a way that does not reveal their underlying private models. The collective objective during inference is for the distributed ensemble of agents to arrive at a shared, globally consistent decision or estimate, i.e., to achieve consensus across the entire grid.

Our proposed architecture is organized as a two-layer hierarchy, where each layer contains a TM, as depicted in Fig. \ref{fig:architecture}. The TM in the first layer (the Input Layer) operates directly on the Boolean input features. During the learning phase, it forms clauses that specialize in recognizing particular patterns present in these input features. 

The TM in the second layer (Neighborhood Aggregation Layer) receives, as its input, the outputs of the clauses from the first-layer TM associated with its own cell, together with the clause outputs originating from the eight nearest neighboring cells. Based on this combined input, the Neighborhood Aggregation Layer TM learns to detect higher-level, more global response patterns emerging from the local cluster comprising the cell itself and its nearest neighbors. In a multi-class classification scenario, each agent maintains a pair of TMs for every class.

As empirically demonstrated in Section \ref{sec:experimentation}, the proposed aggregation layer yields superior performance relative to simpler, non–learned aggregation strategies, such as voting-sum aggregation. In particular, the clause-output states exhibit a stronger and more precise correlation with the clusters’ input features than those produced by the aggregated voting sum. The aggregation-layer TM effectively learns the activity patterns of the clauses within the cluster.

Crucially, only the binary clause outputs are exchanged between neighboring cells, while the internal structure and composition of the clauses remain private and are not revealed. Because each clause output can be encoded as a single bit, the communication overhead between neighbors is kept very low, substantially reducing the volume of data that needs to be transmitted.

\subsection{Distributed Learning}
\label{subsec:learning}

 The complete training workflow is detailed in Algorithm \ref{alg:training algorithm}. The training of an agent is carried out in two distinct stages. In the first stage, the Input Layer TM is trained exclusively from the Boolean feature representation provided as input (lines 6, 7). During the second stage, the Neighborhood Aggregation Layer TM is trained (line 10) while the Input Layer TM, as well as the Input Layer TMs of neighboring agents, are kept in inference mode, thereby supplying their outputs as input signals. The Input Layer clause outputs of the nearest neighbors are gathered (lines 8, 9) via a dedicated communication protocol, the details of which will be presented and analyzed in Section~\ref{subsec: Communication}.

In a multi-class classification scenario, each agent maintains a pair of TMs for every class. For a given training instance, the TM pair (consisting of the Input Layer and the Neighborhood Aggregation TMs) associated with the instance’s ground-truth label is updated first (lines 6-10). After this label-specific update, the training step is repeated for another pair of TMs corresponding to a different class (lines 11-15), which is selected at random but known to all agents. This sequence of operations defines a single training iteration for the agent.

\begin{algorithm}[!h]
\caption{Algorithm: Node's Learning Procedure.}
\label{alg:training algorithm}
\begin{algorithmic}[1]

\Statex \textbf{Notations}
\State \text {$\textit{INP}_i^c$ - Input Layer TM of node $i$, class $c$}
\State \text {$\textit{AGR}_i^c$ - Aggregation Layer TM of node $i$, class $c$}
\State \text {$N$ - Number of classes}
\Statex
\Require    $input\_features, class\_labels, $
\item[] \hspace*{3.5em} $alt\_class\_labels$
\Statex 
     \ForAll {examples \{$input\_features, class\_label$\}}
          \State {Initialize the local record of $\text{INP}_i^c$ clause outputs for c=1,2,..,N and $\forall i \in$ \{node's nearest neighbors\}}
          \Statex {}
          \Statex \hspace{1em} \textbf{Select and train a TM pair according to the class label} 
                \State {update $\textit{INP}_i^{class\_label}$ for current input} 
                \State {calculate $\textit{INP}_i^{class\_label}$ clause outputs}
                \State {broadcast $\textit{INP}_i^{class\_label}$ clause outputs to nearest neighbors}    
                \State {receive and record $\textit{INP}_j^{class\_label}$ clause outputs $\forall j \in$ neighbors(i)}
                \State update $\textit{AGR}_i^{class\_label}$ using acquired input layer clause outputs related to $class\_label$
            \Statex 
            \Statex \hspace{1em} \textbf{Train a TM pair of a different class.  The different class label is known to all agents} 
                \State update $\textit{INP}_i^{alt\_class\_label}$ for current input, $alt\_class\_label  \ne class\_label$
                \State calculate $\textit{INP}_i^{alt\_class\_label}$ clause outputs 
                \State broadcast $\textit{INP}_i^{alt\_class\_label}$ clause outputs to nearest neighbors  
                \State receive and record $\textit{INP}_j^{alt\_class\_label}, \forall j \in$ neighbors(i)
                \State update $\textit{AGR}_i^{alt\_class\_label}$ using acquired input layer clause outputs related to $alt\_class\_label$
                \Statex
   \EndFor

\end{algorithmic}
\end{algorithm}
  
\subsection{Inter-Agent Communication}
\label{subsec: Communication}

Inter-agent communication is employed both to coordinate agents' activities and to transfer necessary data for processing. In the context of distributed edge computing, the design of the underlying communication protocol becomes a key architectural concern, because the energy overhead associated with transmitting and receiving data typically exceeds the energy cost of performing the computations themselves \cite{Zhao02112018}. Our model assumes a wireless (or other shared media) communication setting where a single transmission can be overheard by multiple nodes. 
We employ a gossip random broadcast scheme, akin to that presented in \cite{4787122} (nodes asynchronously broadcast to their neighbors when information is ready). The broadcast mechanism enables an agent to disseminate information to all of its neighboring nodes in a single transmission, thereby accelerating convergence and reducing communication overhead. Furthermore, the use of an asynchronous protocol ensures that each agent operates autonomously, with less dependency on other nodes. This obviates the necessity for a globally synchronized master clock and mitigates delays or bottlenecks associated with synchronization, allowing the system to function in a decentralized and temporally flexible manner.

\subsubsection{Clause Outputs Update}
\label{subsubsec: clause output update} 

After training its Input Layer TM for the given example, the agent calculates the output of the corresponding Input Layer TM clauses. Then it broadcasts these output values to its immediate neighboring agents located within its connectivity radius. Messages are tagged with sample identifiers to ensure that clause outputs correspond to the same training instance. Each clause output can be encoded as a single bit, substantially reducing the volume of data that needs to be transmitted. All agents that successfully receive this communication accordingly update their internal state. Once the originating agent has collected the updated information from each of its nearest neighbors, it initiates the training procedure for its Neighborhood Aggregation Layer TM. However, if any communication packet is lost during this exchange, the agent may omit the Neighborhood Aggregation Layer TM training phase for that cycle.

\subsubsection{Class Vote Sum Update}
\label{subsubsec: vote sum update}
Algorithm \ref{alg:class_vote_sum_update} presents the procedure executed by each agent. For class prediction (inference), each agent's local class vote sums need to be disseminated to all other agents, so that a network-wide consensus on the predicted class can be achieved.  In this context, consensus denotes a state in which every agent possesses complete information regarding the vote sums of every other agent, such that each node can compute an identical, globally consistent class decision. To accomplish this, we apply a multi-source, all-to-all dissemination protocol. 

Each node maintains a vote-sum table that contains one entry for every node in the grid. At the beginning of the procedure, each node transmits its own current vote sum to its eight immediately adjacent neighbors (line 11). When a node receives a message, it updates the corresponding entries in its local vote-sum table with the newly obtained values (lines 12-21).  Furthermore, whenever a node acquires previously unknown vote-sum information, it propagates this new information by broadcasting it to its nearest neighbors (gossip with aggregation) (lines 22-24). A given node can obtain identical data from several independent upstream sources. This redundancy may improve the resilience of the protocol to communication impairments, for example, in the presence of packet loss, intermittent connectivity, or transient link failures, the likelihood that the node still receives the required information remains high. The impact of communication impairments was not evaluated in this study and is deferred for future work. Through repeated local exchanges of this kind, the complete set of vote sums is gradually diffused throughout the grid, enabling all agents to converge on a shared global decision. 
 Assuming reliable delivery and a fair scheduler, all agents become fully informed of all values in $\Theta(n)$ time. The total number of broadcast events is $\Theta(n^2)$ (every agent has to transmit at least once) \cite{2821576, 355459, 1238221}. 

\begin{algorithm}[t]
\caption{Class Vote Sum Update (Executed by Node $i$ for each class)}
\label{alg:class_vote_sum_update}
\begin{algorithmic}[1]

\State \textbf{Local state:}
\State \quad $T[1:n^2]$  \Comment{Vote-sum table}
\State \quad $Known$ \Comment{Set of node IDs with known vote sums}
\State \quad $vote_i$ \Comment{Local vote of node $i$}

\vspace{0.5em}
\State \textbf{Initialization:}
\For{$v = 1$ to $n^2$}
    \State $T[v] \leftarrow null$
\EndFor
\vspace{0.5em}
\State $T[i] \leftarrow vote_i$
\State $Known \leftarrow \{i\}$
\State \textbf{broadcast} $\{(i, vote_i)\}$ to $neighbors(i)$

\vspace{0.5em}
\State \textbf{Upon receiving message $M$:} 
\State \Comment{M contains (node\_id, vote\_sum) tuples}
\State $NewInfo \leftarrow \emptyset$
\ForAll{$(v, s) \in M$}
    \If{$T[v] = null$} 
        \State $T[v] \leftarrow s$
        \State $Known \leftarrow Known \cup \{v\}$
        \State $NewInfo \leftarrow NewInfo \cup \{(v, s)\}$
    \EndIf
\EndFor

\If{$NewInfo \neq \emptyset$}
    \State \textbf{broadcast} $NewInfo$ to $neighbors(i)$
\EndIf

\vspace{0.5em}
\State \textbf{Local completion condition:}
\State Node $i$ is fully informed when $|Known| = n^2$

\end{algorithmic}
\end{algorithm}

\section{Experimentation}
\label{sec:experimentation}

In the initial assessment of our distributed ML model, we operate under the assumption of ideal communication among all agents; consequently, practical network imperfections such as packet loss, delays, or message corruption are not yet incorporated into the analysis. For empirical evaluation, we employ the MNIST dataset of handwritten digits as well as the Fashion-MNIST dataset. 

The distributed computational architecture is organized as an $n \times n$ two-dimensional lattice, yielding a total of $n^2$ agents. Each image from the datasets is spatially partitioned into $n^2$ equally sized tiles, with each tile uniquely assigned to one agent in the grid. All agents share the same TM hyperparameters (number of clauses, threshold $T$, and specificity parameter $s$), ensuring a homogeneous configuration at the architectural level. Nevertheless, each agent's internal model is adapted to the local data it receives and thus learns to detect and classify patterns that are characteristic of the particular image tile for which it is responsible.

\subsection{MNIST Benchmark}
\label{sec: MNIST}

\begin{figure}[!h]
        \centering
         \includegraphics[width=\columnwidth, trim = {0cm 0cm 10.5cm 0cm}, clip]{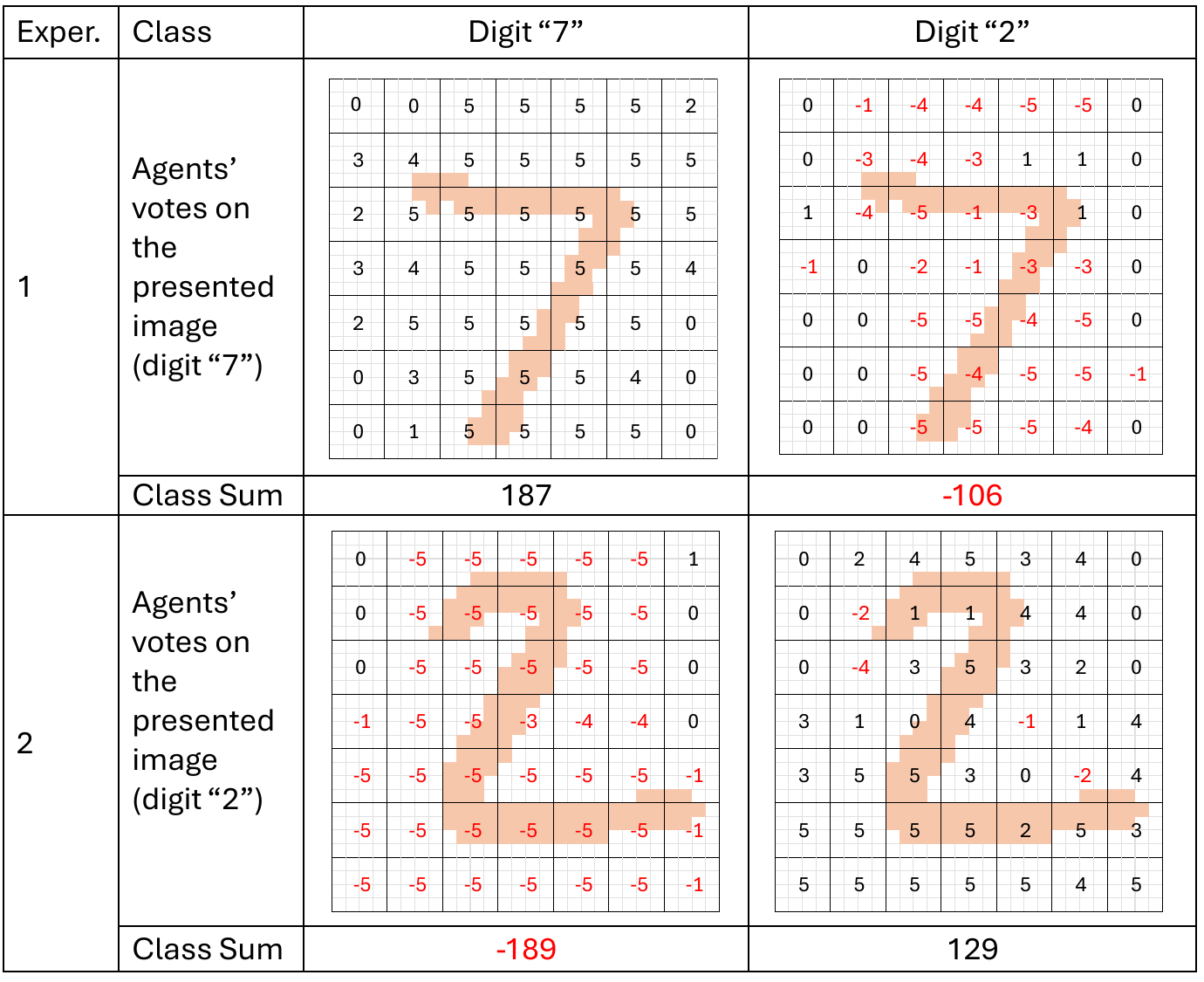}
        \caption{MNIST benchmark: Class vote sums.  The number in each tile is the agent's vote for that tile.}
        \label{fig:MNIST vote sums}
    \end{figure}

The MNIST dataset \cite{lecun2010mnist} is composed of images with a resolution of \(28 \times 28\) pixels. Classification performance on this dataset is reported for two distinct experimental setups. In the first setup, the system is composed of 49 agents, where each agent is responsible for processing a \(4 \times 4\) pixel subregion (tile) of the original image. In this configuration, each tile covers approximately \(2.05\%\) of the entire image area. The collective decision of the ensemble of 49 participating agents is then used to assign a class label to the complete image.
Fig. \ref{fig:MNIST vote sums} presents the vote patterns of the 49 agents associated with the class representing the digit ``7" and of those associated with the class representing the digit ``2" for two images. The number in each tile is the agent's vote for that tile. In the first experiment, an image of a handwritten 7 is provided as input, whereas in the second experiment, the input is an image of a handwritten 2.   As the chart shows, the aggregated class-wise vote sum serves as a robust discriminator, reliably separating the true class from competing classes.

Table \ref{tab:MNIST 49 agents} reports the test accuracies obtained after 50 training epochs by an ensemble comprising 49 agents, where each agent is responsible for learning a $4 \times 4$ image patch. The table reports test accuracy results obtained from single-layer and two-layer agent architectures. The performance of the two-layer architecture is substantially better than that of a single layer. The results are shown for multiple architectural settings of both the Input Layer TM and the Neighborhood Aggregation TM, enabling a comparison of performance across different configuration choices.

\begin{table}[]
\centering
\begin{threeparttable}
\caption{MNIST test accuracy.  49 agents, each learns a $4 \times 4$ image tile. A tile contains 2.05\% of the whole image.}
\label{tab:MNIST 49 agents}
\begin{tabular}{|ccc|ccc|c|c|}
\hline
\multicolumn{3}{|c|}{Input layer} &
  \multicolumn{3}{c|}{Aggregation Layer} &
  \multicolumn{1}{l|}{\multirow{2}{*}{Epochs}} &
  \multirow{2}{*}{\begin{tabular}[c]{@{}c@{}}Accuracy \\ {[}\%{]}\end{tabular}} \\ \cline{1-6}
\multicolumn{1}{|c|}{Clauses} &
  \multicolumn{1}{c|}{T} &
  s &
  \multicolumn{1}{c|}{Clauses} &
  \multicolumn{1}{c|}{T} &
  s &
  \multicolumn{1}{l|}{} &
   \\ \hline
\multicolumn{1}{|c|}{16}  & \multicolumn{1}{c|}{3} & 10 & \multicolumn{3}{c|}{\multirow{4}{*}{vote sums}}             & \multirow{4}{*}{20} & 89.23 \\ \cline{1-3} \cline{8-8} 
\multicolumn{1}{|c|}{32}  & \multicolumn{1}{c|}{4} & 10 & \multicolumn{3}{c|}{}                                 &                     & 89.94 \\ \cline{1-3} \cline{8-8} 
\multicolumn{1}{|c|}{64}  & \multicolumn{1}{c|}{6} & 10 & \multicolumn{3}{c|}{}                                 &                     & 90.80 \\ \cline{1-3} \cline{8-8} 
\multicolumn{1}{|c|}{128} & \multicolumn{1}{c|}{8} & 20 & \multicolumn{3}{c|}{}                                 &                     & 90.86 \\ \hline
\multicolumn{1}{|c|}{16} &
  \multicolumn{1}{c|}{3} &
  3 &
  \multicolumn{1}{c|}{32} &
  \multicolumn{1}{c|}{4} &
  10 &
  \multirow{6}{*}{50} &
  92.63 \\ \cline{1-6} \cline{8-8} 
\multicolumn{1}{|c|}{16}  & \multicolumn{1}{c|}{3} & 3  & \multicolumn{1}{c|}{48} & \multicolumn{1}{c|}{5} & 10 &                     & 92.90 \\ \cline{1-6} \cline{8-8} 
\multicolumn{1}{|c|}{16}  & \multicolumn{1}{c|}{3} & 3  & \multicolumn{1}{c|}{64} & \multicolumn{1}{c|}{4} & 10 &                     & 93.60 \\ \cline{1-6} \cline{8-8} 
\multicolumn{1}{|c|}{32}  & \multicolumn{1}{c|}{3} & 3  & \multicolumn{1}{c|}{32} & \multicolumn{1}{c|}{4} & 10 &                     & 93.21 \\ \cline{1-6} \cline{8-8} 
\multicolumn{1}{|c|}{32}  & \multicolumn{1}{c|}{3} & 3  & \multicolumn{1}{c|}{48} & \multicolumn{1}{c|}{5} & 10 &                     & 93.84 \\ \cline{1-6} \cline{8-8} 
\multicolumn{1}{|c|}{32}  & \multicolumn{1}{c|}{3} & 3  & \multicolumn{1}{c|}{64} & \multicolumn{1}{c|}{6} & 10 &                     & 94.40 \\ \hline
\end{tabular}
\begin{tablenotes}
\footnotesize
\item Accuracy of a classic TM with 200 clauses, T=10, s=7,5, is 96.73\% @100 epochs \cite{pyTsetlinMachine}.
\end{tablenotes}
\end{threeparttable}
\end{table}

Table \ref{tab:MNIST 16 agents} presents the test accuracies obtained after 50 training epochs by an ensemble comprising 16 agents, where each agent is responsible for learning a $7 \times 7$ image tile. In this setup, each agent learns a tile that contains 6.25\% of the whole image. The table reports test accuracy results obtained from single-layer and two-layer agent architectures, showing the superior performance of the two-layer architecture.

\begin{table}[]
\centering
\begin{threeparttable}
\caption{MNIST test accuracy.  16 agents, each learns a $7 \times 7$ image tile. A tile contains 6.25\% of the whole image.}
\label{tab:MNIST 16 agents}
\begin{tabular}{|ccc|ccc|c|c|}
\hline
\multicolumn{3}{|c|}{Input layer} &
  \multicolumn{3}{c|}{Aggregation Layer} &
  \multicolumn{1}{l|}{\multirow{2}{*}{Epochs}} &
  \multirow{2}{*}{\begin{tabular}[c]{@{}c@{}}Accuracy \\ {[}\%{]}\end{tabular}} \\ \cline{1-6}
\multicolumn{1}{|c|}{Clauses} &
  \multicolumn{1}{c|}{T} &
  s &
  \multicolumn{1}{c|}{Clauses} &
  \multicolumn{1}{c|}{T} &
  s &
  \multicolumn{1}{l|}{} &
   \\ \hline
\multicolumn{1}{|c|}{16}  & \multicolumn{1}{c|}{3} & 10 & \multicolumn{3}{c|}{\multirow{4}{*}{vote sums}}             & \multirow{4}{*}{20} & 89.61 \\ \cline{1-3} \cline{8-8} 
\multicolumn{1}{|c|}{32}  & \multicolumn{1}{c|}{4} & 10 & \multicolumn{3}{c|}{}                                 &                     & 91.40 \\ \cline{1-3} \cline{8-8} 
\multicolumn{1}{|c|}{64}  & \multicolumn{1}{c|}{6} & 10 & \multicolumn{3}{c|}{}                                 &                     & 90.80 \\ \cline{1-3} \cline{8-8} 
\multicolumn{1}{|c|}{128} & \multicolumn{1}{c|}{8} & 20 & \multicolumn{3}{c|}{}                                 &                     & 93.63 \\ \hline
\multicolumn{1}{|c|}{16} &
  \multicolumn{1}{c|}{4} &
  5 &
  \multicolumn{1}{c|}{16} &
  \multicolumn{1}{c|}{3} &
  10 &
  \multirow{6}{*}{50} &
  92.65 \\ \cline{1-6} \cline{8-8} 
\multicolumn{1}{|c|}{16}  & \multicolumn{1}{c|}{4} & 5  & \multicolumn{1}{c|}{32} & \multicolumn{1}{c|}{4} & 10 &                     & 93.76 \\ \cline{1-6} \cline{8-8} 
\multicolumn{1}{|c|}{16}  & \multicolumn{1}{c|}{4} & 5  & \multicolumn{1}{c|}{64} & \multicolumn{1}{c|}{6} & 10 &                     & 93.88 \\ \cline{1-6} \cline{8-8} 
\multicolumn{1}{|c|}{32}  & \multicolumn{1}{c|}{4} & 5  & \multicolumn{1}{c|}{32} & \multicolumn{1}{c|}{4} & 10 &                     & 94.12 \\ \cline{1-6} \cline{8-8} 
\multicolumn{1}{|c|}{32}  & \multicolumn{1}{c|}{4} & 5  & \multicolumn{1}{c|}{48} & \multicolumn{1}{c|}{5} & 10 &                     & 94.60 \\ \cline{1-6} \cline{8-8} 
\multicolumn{1}{|c|}{32}  & \multicolumn{1}{c|}{4} & 5  & \multicolumn{1}{c|}{64} & \multicolumn{1}{c|}{6} & 10 &                     & 94.65 \\ \hline
\end{tabular}
\begin{tablenotes}
\footnotesize
\item Accuracy of a classic TM with 200 clauses, T=10, s=7.5, is 96.73\% @100 epochs \cite{pyTsetlinMachine}.
\end{tablenotes}
\end{threeparttable}
\end{table}

\subsection{Fashion-MNIST Benchmark}
\label{subsec:fashion-MNIST}

The Fashion-MNIST dataset \cite{Xiao2017FashionMNISTAN} consists of images with a resolution of $28 \times 28$ pixels, depicting items from 10 distinct categories of apparel (such as T-shirts, trousers, and bags). We begin by analyzing the behavior of an ensemble model composed of 49 agents, where each agent processes a dedicated $4 \times 4$ pixel patch of the input image, corresponding to 2.05\% of the entire image area. 

\begin{figure}[!h]
        \centering
         \includegraphics[width=\columnwidth, trim = {0cm 0cm 5.25cm 0cm}, clip]{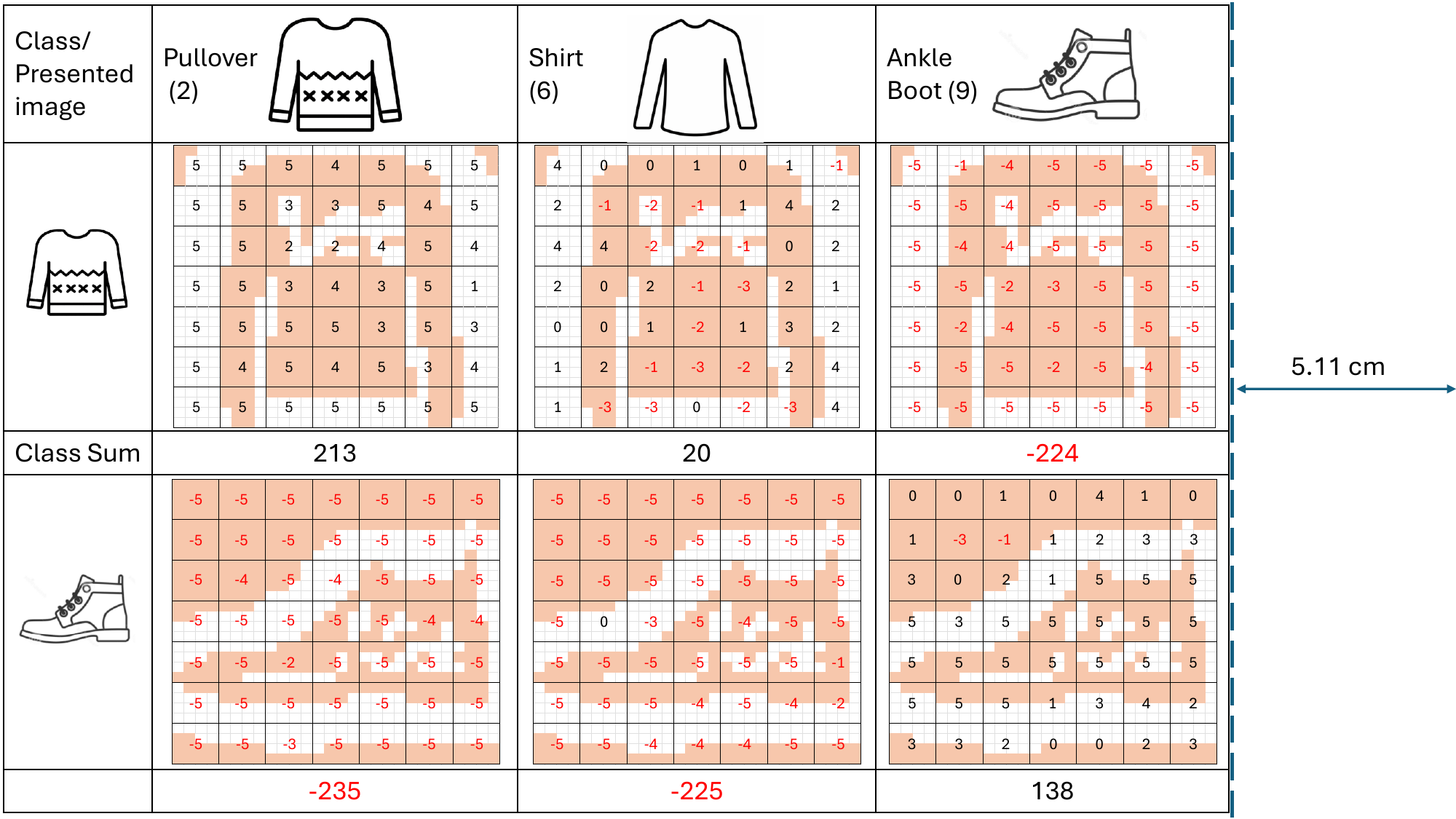}
        \caption{Fashion-MNIST benchmark: agents' voting behavior for 3 classes. The number in each tile is the agent's vote for that tile.}
        \label{fig:Fashion-MNIST vote sums}
    \end{figure}

For the 49-agent Fashion-MNIST setup, Fig. \ref{fig:Fashion-MNIST vote sums} illustrates the voting behavior of the agents for three specific classes (pullover, shirt, and ankle boot), along with the resulting aggregated class scores, for two example input images: a Pullover and an ankle boot. The number in each tile is the agent's vote for that tile. In both examples, the correct class label is clearly identifiable from the overall class sum, which stands out relative to the competing classes. 

For the pullover input, the Shirt class also attains a positive cumulative vote, reflecting the visual similarity between shirts and pullovers in the dataset (e.g., similar silhouettes or overlapping features). Nevertheless, the pullover class still exhibits a substantially higher class sum than the shirt class, indicating that the ensemble is able to correctly disambiguate between these two visually related categories.

Table \ref{F-MNIST 49 agents} summarizes the test accuracies achieved for the Fashion-MNIST dataset after 50 training epochs by an ensemble of 49 agents, with each agent dedicated to modeling a distinct $4 \times 4$ image patch. The table presents results for several architectural parameters of both the Input Layer TM and the Neighborhood Aggregation TM. 

\begin{table}[]
\centering
\begin{threeparttable}
\caption{Fashion-MNIST test accuracy.  49 agents, each learns a $4 \times 4$ tile. A tile contains 2.05\% of the whole image.}
\label{F-MNIST 49 agents}
\begin{tabular}{|ccc|ccc|c|c|}
\hline
\multicolumn{3}{|c|}{Input layer} &
  \multicolumn{3}{c|}{Aggregation Layer} &
  \multicolumn{1}{l|}{\multirow{2}{*}{Epochs}} &
  \multirow{2}{*}{\begin{tabular}[c]{@{}c@{}}Accuracy \\ {[}\%{]}\end{tabular}} \\ \cline{1-6}
\multicolumn{1}{|c|}{Clauses} &
  \multicolumn{1}{c|}{T} &
  s &
  \multicolumn{1}{c|}{Clauses} &
  \multicolumn{1}{c|}{T} &
  s &
  \multicolumn{1}{l|}{} &
   \\ \hline
\multicolumn{1}{|c|}{16}  & \multicolumn{1}{c|}{3} & 10 & \multicolumn{3}{c|}{\multirow{4}{*}{vote sum}}        & \multirow{4}{*}{20} & 74.89 \\ \cline{1-3} \cline{8-8} 
\multicolumn{1}{|c|}{32}  & \multicolumn{1}{c|}{4} & 10 & \multicolumn{3}{c|}{}                                 &                     & 75.69 \\ \cline{1-3} \cline{8-8} 
\multicolumn{1}{|c|}{64}  & \multicolumn{1}{c|}{6} & 10 & \multicolumn{3}{c|}{}                                 &                     & 74.96 \\ \cline{1-3} \cline{8-8} 
\multicolumn{1}{|c|}{128} & \multicolumn{1}{c|}{8} & 10 & \multicolumn{3}{c|}{}                                 &                     & 76.86 \\ \hline
\multicolumn{1}{|c|}{16}  & \multicolumn{1}{c|}{3} & 3  & \multicolumn{1}{c|}{32} & \multicolumn{1}{c|}{4} & 10 & \multirow{5}{*}{50} & 80.75 \\ \cline{1-6} \cline{8-8} 
\multicolumn{1}{|c|}{16}  & \multicolumn{1}{c|}{3} & 3  & \multicolumn{1}{c|}{48} & \multicolumn{1}{c|}{5} & 10 &                     & 81.34 \\ \cline{1-6} \cline{8-8} 
\multicolumn{1}{|c|}{16}  & \multicolumn{1}{c|}{3} & 3  & \multicolumn{1}{c|}{64} & \multicolumn{1}{c|}{6} & 10 &                     & 81.64 \\ \cline{1-6} \cline{8-8} 
\multicolumn{1}{|c|}{32}  & \multicolumn{1}{c|}{3} & 3  & \multicolumn{1}{c|}{32} & \multicolumn{1}{c|}{4} & 10 &                     & 82.11 \\ \cline{1-6} \cline{8-8} 
\multicolumn{1}{|c|}{32}  & \multicolumn{1}{c|}{3} & 3  & \multicolumn{1}{c|}{48} & \multicolumn{1}{c|}{5} & 10 &                     & 82.41 \\ \hline
\multicolumn{1}{|c|}{32}  & \multicolumn{1}{c|}{3} & 3  & \multicolumn{1}{c|}{64} & \multicolumn{1}{c|}{6} & 10 & 35                  & 83.85 \\ \hline
\end{tabular}
\begin{tablenotes}
\footnotesize
\item Accuracy of a classic TM with 500 clauses, T=20, s=7.5, is 87.75\% @100 epochs \cite{pyTsetlinMachine}.
\end{tablenotes}
\end{threeparttable}
\end{table}

Table \ref{F-MNIST 16 agents} summarizes the test accuracies achieved for the Fashion-MNIST dataset after 50 training epochs by an ensemble of 16 agents, with each agent dedicated to modeling a distinct $7 \times 7$ image patch. The table presents results for several architectural parameters of both the Input Layer TM and the Neighborhood Aggregation TM. 

\begin{table}[]
\centering
\begin{threeparttable}
\caption{Fashion-MNIST test accuracy.  16 agents, each learns a $7 \times 7$ image tile. A tile contains 2.05\% of the whole image.}
\label{F-MNIST 16 agents}
\begin{tabular}{|ccc|ccc|c|c|}
\hline
\multicolumn{3}{|c|}{Input layer} &
  \multicolumn{3}{c|}{Aggregation Layer} &
  \multicolumn{1}{l|}{\multirow{2}{*}{Epochs}} &
  \multirow{2}{*}{\begin{tabular}[c]{@{}c@{}}Accuracy \\ {[}\%{]}\end{tabular}} \\ \cline{1-6}
\multicolumn{1}{|c|}{Clauses} &
  \multicolumn{1}{c|}{T} &
  s &
  \multicolumn{1}{c|}{Clauses} &
  \multicolumn{1}{c|}{T} &
  s &
  \multicolumn{1}{l|}{} &
   \\ \hline
\multicolumn{1}{|c|}{16}  & \multicolumn{1}{c|}{3} & 10 & \multicolumn{3}{c|}{\multirow{4}{*}{vote sum}}        & \multirow{4}{*}{20} & 79.11 \\ \cline{1-3} \cline{8-8} 
\multicolumn{1}{|c|}{32}  & \multicolumn{1}{c|}{4} & 10 & \multicolumn{3}{c|}{}                                 &                     & 80.45 \\ \cline{1-3} \cline{8-8} 
\multicolumn{1}{|c|}{64}  & \multicolumn{1}{c|}{6} & 10 & \multicolumn{3}{c|}{}                                 &                     & 81.02 \\ \cline{1-3} \cline{8-8} 
\multicolumn{1}{|c|}{128} & \multicolumn{1}{c|}{8} & 20 & \multicolumn{3}{c|}{}                                 &                     & 82.94 \\ \hline
\multicolumn{1}{|c|}{16}  & \multicolumn{1}{c|}{3} & 3  & \multicolumn{1}{c|}{16} & \multicolumn{1}{c|}{3} & 10 & \multirow{6}{*}{50} & 80.24 \\ \cline{1-6} \cline{8-8} 
\multicolumn{1}{|c|}{16}  & \multicolumn{1}{c|}{3} & 3  & \multicolumn{1}{c|}{32} & \multicolumn{1}{c|}{4} & 10 &                     & 81.25 \\ \cline{1-6} \cline{8-8} 
\multicolumn{1}{|c|}{16}  & \multicolumn{1}{c|}{3} & 3  & \multicolumn{1}{c|}{64} & \multicolumn{1}{c|}{6} & 10 &                     & 82.96 \\ \cline{1-6} \cline{8-8} 
\multicolumn{1}{|c|}{32}  & \multicolumn{1}{c|}{3} & 3  & \multicolumn{1}{c|}{32} & \multicolumn{1}{c|}{4} & 10 &                     & 83.05 \\ \cline{1-6} \cline{8-8} 
\multicolumn{1}{|c|}{32}  & \multicolumn{1}{c|}{3} & 3  & \multicolumn{1}{c|}{48} & \multicolumn{1}{c|}{5} & 10 &                     & 83.63 \\ \cline{1-6} \cline{8-8} 
\multicolumn{1}{|c|}{32}  & \multicolumn{1}{c|}{3} & 3  & \multicolumn{1}{c|}{64} & \multicolumn{1}{c|}{6} & 10 &                     & 83.99 \\ \hline
\end{tabular}
\begin{tablenotes}
\footnotesize
\item Accuracy of a classic TM with 500 clauses, T=20, s=7.5, is 87.75\% @100 epochs \cite{pyTsetlinMachine}.
\end{tablenotes}
\end{threeparttable}
\end{table}

\subsection{Sensor Network}
\label{subsec:sensor-net}
In this experiment, we model distributed learning across a sensor network with heterogeneity in sensor data.

The setup employs 25 agents arranged in a $5 \times 5$ two-dimensional lattice.  The class semantics (3 classes) are common to all agents; however, the mapping of sensor data to those classes differs across clients.   \textbf{Feature distribution heterogeneity} is applied by:
\begin{itemize}
    \item Agents having one or two sensors (structural heterogeneity).
    \item Sensors having different scaling depending on the agent id (sensor 0: $\times1.8/1.0$, sensor 1: $\times0.6/1$)
    \item Sensor 0 having different offset depending on agent id ($1.2/none$)
    \item Different noise levels on sensor 0 and 1 (zero mean, $std\ 0.3/0.4/0.8$)
\end{itemize}

\begin{figure*}[!h]
        \centering
         \includegraphics[width=16cm]{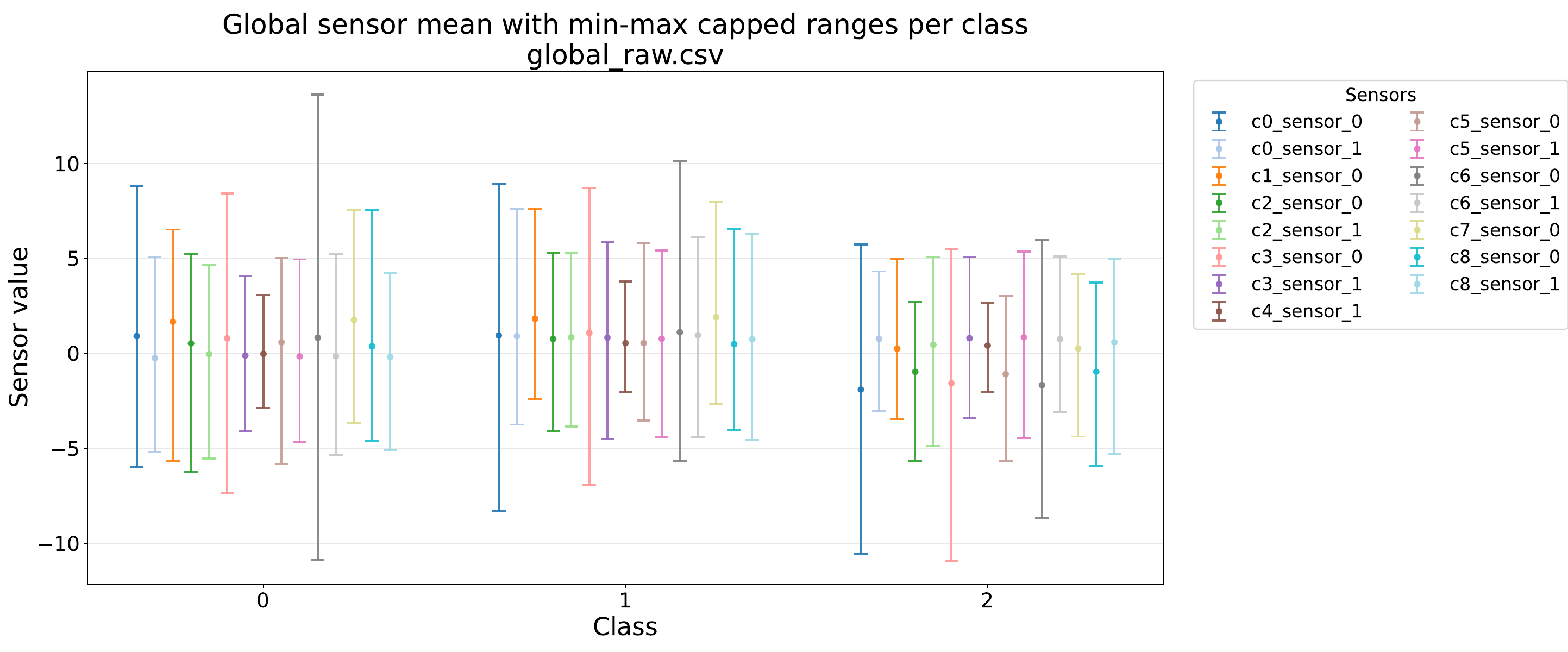}
        \caption{Sensor data distribution for 9 agents.}
        \label{fig:sensor_data_dist}
    \end{figure*}

For each class, the dataset is generated by sampling a base mean uniformly from 
$[-1,1]$. The variance parameters are independently sampled from $[2,3]$, and the resulting values are used to generate samples from a multivariate Gaussian distribution. A sensor-heterogeneity model is then applied to introduce agent-specific adaptations and capture inter-agent variability in the sensing process.  Fig. \ref{fig:sensor_data_dist} presents the distribution of sensor data for 9 agents.

\subsubsection{Classification by ANN}    

Due to the overlap between classes, per-agent classification yields low accuracy. Table \ref{tab:sensor_network} presents the performance obtained using ANN to classify agent2 data samples.  The per-agent accuracy ranged from 43\% to 57.5\%. The architecture of the ANN used is:
\begin{center}
$Input(2) \rightarrow Dense(128, ReLU) \rightarrow Dense(64, Relu) \rightarrow Output(3, Softmax)$   
\end{center}

Since each agent is just a relatively small, noisy sample from that global distribution, training on a single agent's sample set cannot recover the full decision boundary.
The aggregation of samples from all 25 agents was then used to train an ANN model with the following architecture:
\begin{center}
$Input(45) \rightarrow Dense(256, ReLU) \rightarrow Dense(128, Relu) \rightarrow Dense(64, ReLU) \rightarrow Output(3, Softmax)$   
\end{center}
The combined model learns a population-level rule across agents with an accuracy of 93\%. The significant difference between local and global separability is presented in Table \ref{tab:sensor_network}.

\subsubsection{Classification by TM Ensemble}


The sensor measurements are inherently analog quantities that can be modeled as real-valued variables. Since the TM operates on Boolean-valued input features, these continuous variables must first be transformed into a discrete binary representation.

To achieve this, the continuous value range of each feature is partitioned into a predefined number of bins. The binarization strategy adopted in this study determines the bin thresholds via an order-statistic–based procedure, implemented as approximately evenly spaced selections from the sorted list of unique observed values. Concretely, all unique feature values are first extracted and sorted; the smallest unique value is then discarded, and the remaining unique values serve as threshold candidates. From this ordered set, thresholds are chosen at approximately uniform index intervals, yielding an approximation to Quantile Binning \cite{9923830}.

Under ideal Quantile Binning, the continuous data range is partitioned into a given number of bins such that each bin contains approximately the same number of samples. Relative to binning schemes based on fixed, a priori thresholds (e.g., uniformly spaced in the value domain), quantile-based binning is generally more robust for skewed or otherwise non-uniform data distributions, as it promotes a more balanced allocation of samples across bins.

Each feature value is subsequently encoded in a cumulative (monotone) binary fashion with respect to the bin thresholds. For example, if the value range is divided into four bins defined by three thresholds,

\begin{center}
     $t1 < t2 < t3$ ,
\end{center}

the feature value $x$ is encoded in three bits:

\begin{center}
$(x \geq t1)$, $(x \geq t2)$, $(x \geq t3)$ .
\end{center} 

Each sensor sample is binarized to 8 bits (for a two-sensor agent) or 16 bits (for a single sensor agent).

\begin{figure}[!h]
        \centering
         \includegraphics[width=\columnwidth, trim = {0cm 3.1cm 18,2cm 0cm}, clip]{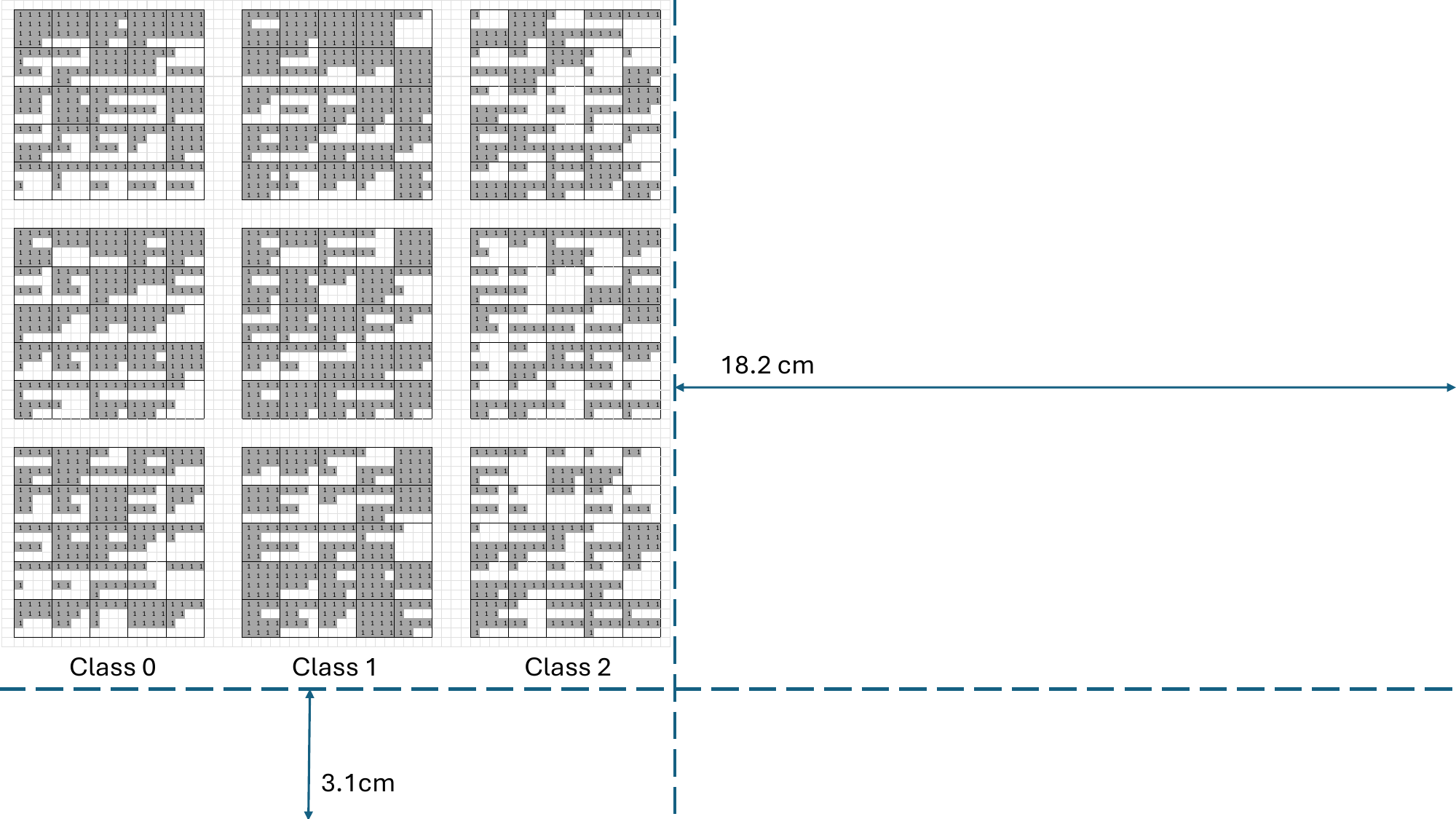}
        \caption{Visualization of the binarized input featues.  3 random frames for each class.}
        \label{fig:feature_visualization}
    \end{figure}

\paragraph{A grid topology}
The 16-bit input feature vector for each agent is first reshaped into a \(4 \times 4\) tile. A total of 25 such tiles (one per agent) are then arranged in a \(5 \times 5\) two-dimensional lattice. Fig. \ref{fig:feature_visualization} shows several randomly selected examples of the resulting \(20 \times 20\) patterns, each representing the global feature set (all 25 agents).  When data from all agents are pooled, a global class structure emerges in the 2D feature space.

\begin{figure}[!h]
        \centering
         \includegraphics[width=\columnwidth, trim = {0cm 8.4cm 16,2cm 0cm}, clip]{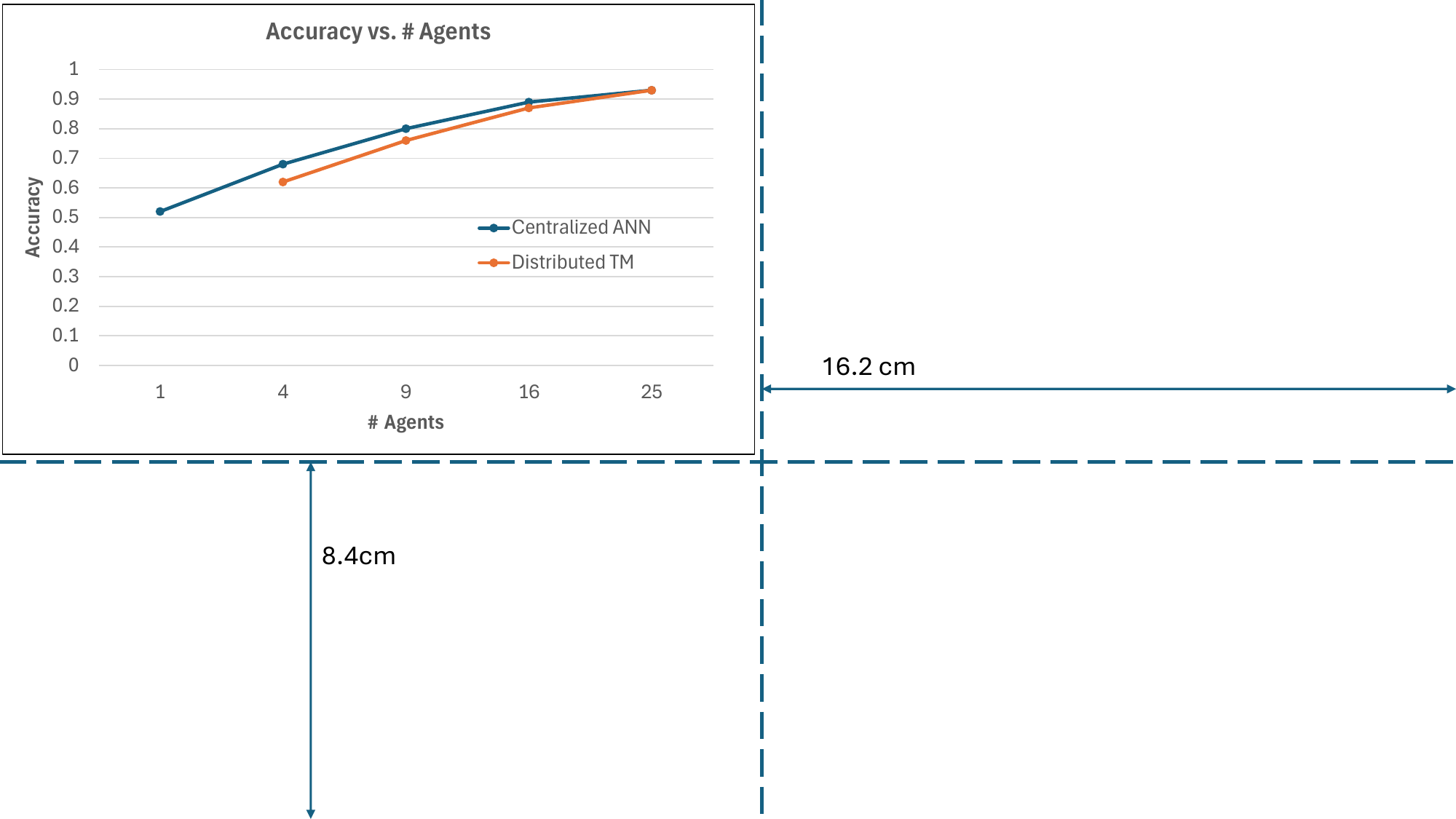}
        \caption{Classification accuracy as a function of the number of participating agents.}
        \label{fig:accuracy_vs_agents}
    \end{figure}

In our setup, each of the 25 agents employs an Input Layer TM with 32 clauses and hyperparameters T=3, s=3.  The Aggregation Layer TM  is of 16 clauses and hyperparameters T=4, s=10. As reported in Table \ref{tab:sensor_network}, the TM ensemble achieves 93\% accuracy, the same level of accuracy achieved by the centralized ANN model. 
The classifier leverages the global class-conditional structure across agents, which yields substantially higher accuracy on the pooled dataset than on individual agents. The chart in Fig. \ref{fig:accuracy_vs_agents} shows classification accuracy as a function of the number of participating agents for both the centralized ANN and distributed TM models. As the number of participating agents increases, the accuracy of the distributed TM progressively converges toward that of the centralized ANN.

\begin{table*}[!h]
\centering
\caption{25-agent sensor network.  Comparison between centralized ANN and collaborative TM learning.}
\label{tab:sensor_network}
\begin{tabular}{|l|ccc|ccc|ccc|}
\hline
\multicolumn{1}{|c|}{\textbf{Model}} &
  \multicolumn{3}{c|}{\textbf{\begin{tabular}[c]{@{}c@{}}ANN\\ Single agent\\ (agent 2)\end{tabular}}} &
  \multicolumn{3}{c|}{\textbf{\begin{tabular}[c]{@{}c@{}}ANN\\ Aggregated data \\ from 25 agents\end{tabular}}} &
  \multicolumn{3}{c|}{\textbf{\begin{tabular}[c]{@{}c@{}}TM\\ Collaborated Learning\\ of 25 agents\end{tabular}}} \\ \hline
Class &
  \multicolumn{1}{c|}{0} &
  \multicolumn{1}{c|}{1} &
  2 &
  \multicolumn{1}{c|}{0} &
  \multicolumn{1}{c|}{1} &
  2 &
  \multicolumn{1}{c|}{0} &
  \multicolumn{1}{c|}{1} &
  2 \\ \hline
Support &
  \multicolumn{1}{c|}{80} &
  \multicolumn{1}{c|}{80} &
  80 &
  \multicolumn{1}{c|}{2000} &
  \multicolumn{1}{c|}{2000} &
  2000 &
  \multicolumn{1}{c|}{2000} &
  \multicolumn{1}{c|}{2000} &
  2000 \\ \hline
Accuracy &
  \multicolumn{3}{c|}{0.50} &
  \multicolumn{3}{c|}{0.93} &
  \multicolumn{3}{c|}{0.93} \\ \hline
Precision &
  \multicolumn{1}{c|}{0.49} &
  \multicolumn{1}{c|}{0.51} &
  0.51 &
  \multicolumn{1}{c|}{0.90} &
  \multicolumn{1}{c|}{0.90} &
  0.99 &
  \multicolumn{1}{c|}{0.92} &
  \multicolumn{1}{c|}{0.89} &
  0.99 \\ \hline
Recall &
  \multicolumn{1}{c|}{0.31} &
  \multicolumn{1}{c|}{0.46} &
  0.74 &
  \multicolumn{1}{c|}{0.90} &
  \multicolumn{1}{c|}{0.90} &
  0.99 &
  \multicolumn{1}{c|}{0.89} &
  \multicolumn{1}{c|}{0.92} &
  0.99 \\ \hline
F1 &
  \multicolumn{1}{c|}{0.38} &
  \multicolumn{1}{c|}{0.48} &
  60 &
  \multicolumn{1}{c|}{0.90} &
  \multicolumn{1}{c|}{0.90} &
  0.99 &
  \multicolumn{1}{c|}{0.90} &
  \multicolumn{1}{c|}{0.90} &
  0.99 \\ \hline
\multirow{3}{*}{\begin{tabular}[c]{@{}l@{}}Confusion\\ Matrix\end{tabular}} &
  \multicolumn{1}{c|}{25} &
  \multicolumn{1}{c|}{25} &
  30 &
  \multicolumn{1}{c|}{1801} &
  \multicolumn{1}{c|}{194} &
  5 &
  \multicolumn{1}{c|}{1773} &
  \multicolumn{1}{c|}{218} &
  9 \\ \cline{2-10} 
 &
  \multicolumn{1}{c|}{16} &
  \multicolumn{1}{c|}{37} &
  27 &
  \multicolumn{1}{c|}{191} &
  \multicolumn{1}{c|}{1802} &
  7 &
  \multicolumn{1}{c|}{152} &
  \multicolumn{1}{c|}{1840} &
  8 \\ \cline{2-10} 
 &
  \multicolumn{1}{c|}{10} &
  \multicolumn{1}{c|}{11} &
  59 &
  \multicolumn{1}{c|}{7} &
  \multicolumn{1}{c|}{15} &
  1978 &
  \multicolumn{1}{c|}{6} &
  \multicolumn{1}{c|}{12} &
  1982 \\ \hline
\end{tabular}
\end{table*}

Table \ref{tab:per_agent_resources} presents the resources used by each agent and the overall system in comparison to a baseline single TM. While each local TM is relatively small, the total number of clauses and automata in the entire system exceeds that of a single traditional TM baseline. Nevertheless, a system made up of multiple small agents can achieve the same classification accuracy by utilizing their private, non-overlapping local samples.

\begin{table*}[h]
\centering
\caption{Per-agent and system-wide TM resources for the sensor network example.  }
\label{tab:per_agent_resources}
\begin{tabular}{|l|c|c|c|c|c|c|c|c|}
\hline
\textbf{} &
  \textbf{Clauses} &
  \textbf{T} &
  \textbf{s} &
  \textbf{Literals} &
  \textbf{TAs} &
  \textbf{\begin{tabular}[c]{@{}c@{}}System-wide \\ instances per class\end{tabular}} &
  \textbf{\begin{tabular}[c]{@{}c@{}}System-wide \\ \# TAs per class\end{tabular}} &
  \textbf{\begin{tabular}[c]{@{}c@{}}Accuracy\\ @ 50 epochs\end{tabular}} \\ \hline
Classic TM Baseline  & 200 & 10 & 5  & 25*16*2 = 800 & 160,000 & 1                   & 160,000                  & 0.93                  \\ \hline
Input Layer TM       & 32  & 3  & 3  & 16*2 = 32     & 1,024   & \multirow{2}{*}{25} & \multirow{2}{*}{256,000} & \multirow{2}{*}{0.93} \\ \cline{1-6}
Aggregation Layer TM & 16  & 4  & 10 & 9*32*2 = 576  & 9,216   &                     &                          &                       \\ \hline
\end{tabular}
\end{table*}

\paragraph{Connected graph topology}

\begin{figure}[!h]
        \centering
         \includegraphics[width=\columnwidth, trim = {0cm 9.2cm 17.3cm 0cm}, clip]{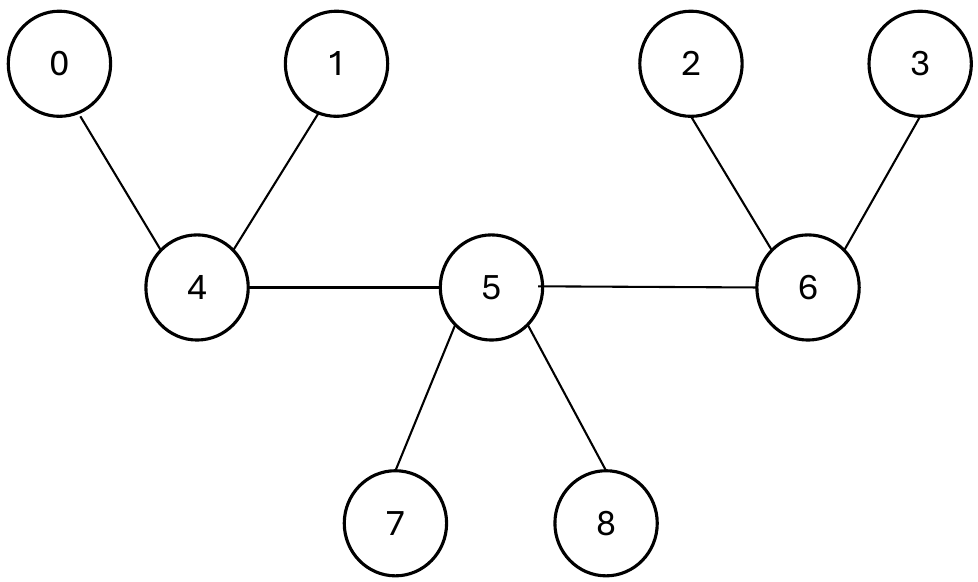}
        \caption{Sensor Network topology graph. Agents with different numbers of nearest neighbors.}
        \label{fig:network_graph}
    \end{figure}

To assess the performance of the algorithm on a more general graph topology, as opposed to the regular $n \times n$ two-dimensional lattice, we consider the network topology illustrated in Fig.~\ref{fig:network_graph}. In this setting, agents possess one, three, or four immediate neighbors, implying that Aggregation Layer TMs associated with different agents operate over feature sets of varying cardinalities, thereby inducing heterogeneous model architectures across the network. As reported in Table \ref{tab:9_agent_network}, the obtained accuracy of 76\% is comparable to that achieved by both the centralized ANN model and the two-dimensional configuration comprising 9 participating agents (see also Fig. \ref{fig:accuracy_vs_agents}).

\begin{table}[]
\centering
\caption{9-agent sensor network as in Fig \ref{fig:network_graph}. Comparison between centralized ANN and collaborative TM learning.}
\label{tab:9_agent_network}
\begin{tabular}{|l|ccc|ccc|}
\hline
\multicolumn{1}{|c|}{\textbf{Model}} &
  \multicolumn{3}{c|}{\textbf{\begin{tabular}[c]{@{}c@{}}ANN\\ Aggregated data \\ from 9 agents\end{tabular}}} &
  \multicolumn{3}{c|}{\textbf{\begin{tabular}[c]{@{}c@{}}TM\\ Collaborated Learning\\ of 9 agents\end{tabular}}} \\ \hline
Class     & \multicolumn{1}{c|}{0}    & \multicolumn{1}{c|}{1}    & 2    & \multicolumn{1}{c|}{0}    & \multicolumn{1}{c|}{1}    & 2    \\ \hline
Support   & \multicolumn{1}{c|}{2000} & \multicolumn{1}{c|}{2000} & 2000 & \multicolumn{1}{c|}{2000} & \multicolumn{1}{c|}{2000} & 2000 \\ \hline
Accuracy  & \multicolumn{3}{c|}{0.80}                                    & \multicolumn{3}{c|}{0.76}                                    \\ \hline
Precision & \multicolumn{1}{c|}{0.75} & \multicolumn{1}{c|}{0.75} & 0.89 & \multicolumn{1}{c|}{0.77} & \multicolumn{1}{c|}{0.67} & 0.86 \\ \hline
Recall    & \multicolumn{1}{c|}{0.75} & \multicolumn{1}{c|}{0.71} & 0.93 & \multicolumn{1}{c|}{0.63} & \multicolumn{1}{c|}{0.77} & 0.89 \\ \hline
F1        & \multicolumn{1}{c|}{0.75} & \multicolumn{1}{c|}{0.73} & 0.91 & \multicolumn{1}{c|}{0.69} & \multicolumn{1}{c|}{0.72} & 0.87 \\ \hline
\multirow{3}{*}{\begin{tabular}[c]{@{}l@{}}Confusion\\ Matrix\end{tabular}} &
  \multicolumn{1}{c|}{1492} &
  \multicolumn{1}{c|}{398} &
  110 &
  \multicolumn{1}{c|}{1252} &
  \multicolumn{1}{c|}{615} &
  133 \\ \cline{2-7} 
          & \multicolumn{1}{c|}{453}  & \multicolumn{1}{c|}{1427} & 120  & \multicolumn{1}{c|}{304}  & \multicolumn{1}{c|}{1545} & 151  \\ \cline{2-7} 
          & \multicolumn{1}{c|}{50}   & \multicolumn{1}{c|}{87}   & 1863 & \multicolumn{1}{c|}{73}   & \multicolumn{1}{c|}{155}  & 1772 \\ \hline
\end{tabular}
\end{table}   

\section {Summary}
\label{sec:Summary}

We propose a hierarchical TM-based framework that supports distributed and decentralized learning, coupled with consensus-oriented inference procedures. The model assumes that training samples are vertically aligned and that the class labels are available to all agents. The framework is explicitly constructed to safeguard the privacy of both the model’s internal parameters and each participant’s locally held data. It naturally accommodates heterogeneous TM-based agents that may differ in their sensing modalities, data acquisition processes, underlying data distributions, or available computational power, thus enabling flexible integration and fusion of information in complex settings such as multi-modal sensing scenarios. During inference, the individual predictions produced by each agent’s local model are aggregated to form a shared, system-wide consensus estimate. Throughout this process, communication is restricted to exchanges between an agent and its immediate neighbors in the network topology, reinforcing both scalability and privacy. 
Experiments were conducted using the MNIST, Fashion-MNIST, and Sensor Network datasets, employing two-dimensional grid and connected graph network topologies. The classification accuracies achieved in these experiments are comparable to those of centralized models. Additionally, the models used by the agents are relatively lightweight, highlighting the potential for edge computing.

In our experimental study, we assume ideal (i.e., error-free and latency-free) communication channels. A more comprehensive evaluation under realistic communication conditions and deployment environments is left for future work.

\section{Data Availability Statement}
The data that support the findings of this study are available from the corresponding author, Y. R., upon reasonable request.


\bibliographystyle{IEEEtran}
\balance

\bibliography{bibliography/article_bibliography,bibliography/abbreviations}

@STRING{SENSORS           = "{IEEE} Sensors J."}

@STRING{Sensors         = "Proc. of {IEEE} Sensors"}

@article{Granmo2018TheTM,
  title={The Tsetlin Machine - A Game Theoretic Bandit Driven Approach to Optimal Pattern Recognition with Propositional Logic},
  author={Ole-Christoffer Granmo},
  year={2018},
 journal={arXiv preprint arXiv:1804.01508},
}

@article{lei2021lowpower,
      title={Low-Power Audio Keyword Spotting using {Tsetlin} Machines}, 
      author={Jie Lei and Tousif Rahman and Rishad Shafik and Adrian Wheeldon and Alex Yakovlev and Ole-Christoffer Granmo and Fahim Kawsar and Akhil Mathur},
      year={2021},
     journal={arXiv preprint arXiv:2101.11336},
}

@inproceedings{10.1145/3560905.3568512,
author = {Bakar, Abu and Rahman, Tousif and Shafik, Rishad and Kawsar, Fahim and Montanari, Alessandro},
title = {Adaptive Intelligence for Batteryless Sensors Using Software-Accelerated {Tsetlin} Machines},
year = {2023},
isbn = {9781450398862},
publisher = {Association for Computing Machinery},
address = {New York, NY, USA},
url = {https://doi.org/10.1145/3560905.3568512},
doi = {10.1145/3560905.3568512},
booktitle = {Proceedings of the 20th ACM Conference on Embedded Networked Sensor Systems},
pages = {236–249},
numpages = {14},
location = {Boston, Massachusetts},
series = {SenSys '22}
}

@article{bhattarai2023tsetlin,
      title={Tsetlin Machine Embedding: Representing Words Using Logical Expressions}, 
      author={Bimal Bhattarai and Ole-Christoffer Granmo and Lei Jiao and Rohan Yadav and Jivitesh Sharma},
      year={2023},
      journal={arXiv preprint arXiv:2301.00709},
}

@article{https://doi.org/10.1111/exsy.12873,
author = {Saha, Rupsa and Granmo, Ole-Christoffer and Goodwin, Morten},
title = {Using {Tsetlin} Machine to discover interpretable rules in natural language processing applications},
journal = {Expert Systems},
volume = {40},
number = {4},
pages = {e12873},
doi = {https://doi.org/10.1111/exsy.12873},
url = {https://onlinelibrary.wiley.com/doi/abs/10.1111/exsy.12873},
eprint = {https://onlinelibrary.wiley.com/doi/pdf/10.1111/exsy.12873},
year = {2023}
}

@article{bhattarai2020measuring,
      title={Measuring the Novelty of Natural Language Text Using the Conjunctive Clauses of a {Tsetlin} Machine Text Classifier}, 
      author={Bimal Bhattarai and Ole-Christoffer Granmo and Lei Jiao},
      year={2020},
      journal={arXiv preprint arXiv:2011.08755},
}

@article{Yadav_Jiao_Granmo_Goodwin_2021,
title={Human-Level Interpretable Learning for Aspect-Based Sentiment Analysis}, volume={35}, 
url={https://ojs.aaai.org/index.php/AAAI/article/view/17671}, 
DOI={10.1609/aaai.v35i16.17671}, 
abstractNote={This paper proposes human-interpretable learning of aspect-based sentiment analysis (ABSA), employing the recently introduced Tsetlin Machines (TMs). We attain interpretability by converting the intricate position-dependent textual semantics into binary form, mapping all the features into bag-of-words (BOWs). The binary form BOWs are encoded so that the information on the aspect and context words are nearly lossless for sentiment classification. We further adapt the BOWs as input to the TM, enabling learning of aspect-based sentiment patterns in propositional logic. To evaluate interpretability and accuracy, we conducted experiments on two widely used ABSA datasets of SemEval 2014: Restaurant 14 and Laptop 14. The experiments show how each relevant feature takes part in conjunctive clauses that contain the context information for the corresponding aspect word, demonstrating human-level interpretability. At the same time, the obtained accuracy is competitive with existing neural network models, reaching 78.02% on Restaurant 14 and 73.51% on Laptop 14.}, 
number={16}, 
journal={Proceedings of the AAAI Conference on Artificial Intelligence}, 
author={Yadav, Rohan K and Jiao, Lei and Granmo, Ole-Christoffer and Goodwin, Morten}, year={2021}, 
month={May}, 
pages={14203-14212} 
}

@misc{granmo2019convolutionaltsetlinmachine,
      title={The Convolutional Tsetlin Machine}, 
      author={Ole-Christoffer Granmo and Sondre Glimsdal and Lei Jiao and Morten Goodwin and Christian W. Omlin and Geir Thore Berge},
      year={2019},
      eprint={1905.09688},
      archivePrefix={arXiv},
      primaryClass={cs.LG},
      url={https://arxiv.org/abs/1905.09688}, 
}

@inproceedings{granmo2019ctm,
  title={Convolutional Tsetlin Machine: Learning interpretable representations through convolutional clauses},
  author={Granmo, Ole-Christoffer and Jiao, Lei},
  booktitle={International Joint Conference on Neural Networks (IJCNN)},
  year={2019}
}

@article{Wheeldon2020LearningAB,
  title={Learning automata based energy-efficient {AI} hardware design for {IoT} applications},
  author={Adrian Wheeldon and Rishad A. Shafik and Tousif and Rahman and Jie Lei and Alex Yakovlev and Ole-Christoffer Granmo},
  journal={Philosophical transactions. Series A, Mathematical, physical, and engineering sciences},
  year={2020},
  volume={378},
  url={https://api.semanticscholar.org/CorpusID:221244332}
}

@INPROCEEDINGS{10455063,
  author={Gunvaldsen, Ole and Thorsen, Henning Blomfeldt and Andersen, Per-Arne and Granmo, Ole-Christoffer and Goodwin, Morten},
  booktitle={2023 International Symposium on the Tsetlin Machine (ISTM)}, 
  title={Towards IoT Anomaly Detection with Tsetlin Machines}, 
  year={2023},
  volume={},
  number={},
  pages={1-8},
  doi={10.1109/ISTM58889.2023.10455063}}

@book{1974automation,
  title={Automation Theory and Modelling of Biological Systems},
  author={M. L. Tsetlin},
  isbn={9780080956114},
  series={Mathematics in Science and Engineering},
  url={https://books.google.co.il/books?id=3wLEDm\_\_bnsC},
  year={1974},
  publisher={Academic Press}
}

@article{M.L.Tsetlin_1963,
doi = {10.1070/RM1963v018n04ABEH001139},
url = {https://dx.doi.org/10.1070/RM1963v018n04ABEH001139},
year = {1963},
month = {aug},
publisher = {},
volume = {18},
number = {4},
pages = {1},
author = {M L Tsetlin},
title = {FINITE AUTOMATA AND MODELS OF SIMPLE FORMS OF BEHAVIOUR},
journal = {Russian Mathematical Surveys}
}

@book{PuppetsWithoutStrings,
  title={Puppets Without Strings},
  author={V. I. Varshavsky, D. A. Pospelov},
  year={1988},
  publisher={MIR}
}

@article{swarm_intelligence,
author = {Garnier, Simon and Gautrais, Jacques and Theraulaz, Guy},
year = {2007},
month = {10},
pages = {3-31},
title = {The biological principles of swarm intelligence},
volume = {1},
journal = {Swarm Intelligence},
doi = {10.1007/s11721-007-0004-y}
}

@ARTICLE{10.3389/frai.2025.1377944,
AUTHOR={Elmisadr, Negar  and Belaid, Mohamed-Bachir  and Yazidi, Anis },     
TITLE={Stochastic and deterministic processes in Asymmetric Tsetlin Machine}, 
JOURNAL={Frontiers in Artificial Intelligence},   
VOLUME={Volume 8 - 2025},
YEAR={2025},
URL={https://www.frontiersin.org/journals/artificial-intelligence/articles/10.3389/frai.2025.1377944},
DOI={10.3389/frai.2025.1377944},
ISSN={2624-8212}}

@inproceedings{Kuruge_2024,
   title={The Probabilistic Tsetlin Machine: A Novel Approach to Uncertainty Quantification},
   url={http://dx.doi.org/10.1145/3704137.3704143},
   DOI={10.1145/3704137.3704143},
   booktitle={Proceedings of the 2024 8th International Conference on Advances in Artificial Intelligence},
   publisher={ACM},
   author={Kuruge, Darshana Abeyrathna and El Mekkaoui, Sara and Hafver, Andreas and Agrell, Christian},
   year={2024},
   month=oct, pages={39–47},
   collection={ICAAI 2024} }

@article{abeyrathna2019regression,
  title={The Regression Tsetlin Machine: A Tsetlin Machine for Continuous Output Problems},
  author={Abeyrathna, K. Darshana and Granmo, Ole-Christoffer and Jiao, Lei and Goodwin, Matthew},
  journal={arXiv preprint arXiv:1905.04206},
  year={2019}
}

@article{jiao2020mctm,
  title={Multi-Class Tsetlin Machine},
  author={Jiao, Lei and Abeyrathna, K. Darshana and Granmo, Ole-Christoffer},
  journal={arXiv preprint arXiv:2001.08227},
  year={2020}
}

@inproceedings{abeyrathna2020weighted,
  title={Weighted Tsetlin Machines: Towards interpretable accurate low-complexity classifiers},
  author={Abeyrathna, K. Darshana and Granmo, Ole-Christoffer and Jiao, Lei and Goodwin, Matthew},
  booktitle={IEEE Symposium Series on Computational Intelligence (SSCI)},
  year={2020}
}

@article{abeyrathna2021iwtm,
  title={Integer Weighted Tsetlin Machines},
  author={Abeyrathna, K. Darshana and Granmo, Ole-Christoffer and Goodwin, Matthew},
  journal={arXiv preprint arXiv:2101.10347},
  year={2021}
}

@inproceedings{abeyrathna2021cmtm,
  title={Coalesced Multi-Output Tsetlin Machine},
  author={Abeyrathna, K. Darshana and Jiao, Lei and Granmo, Ole-Christoffer},
  booktitle={International Joint Conference on Neural Networks (IJCNN)},
  year={2021}
}

@article{yadav2021continuous,
  title={Continuous Input Tsetlin Machine},
  author={Yadav, Vikas and Granmo, Ole-Christoffer and Jiao, Lei},
  journal={arXiv preprint arXiv:2101.10346},
  year={2021}
}

@article{abeyrathna2021recurrent,
  title={The Recurrent Tsetlin Machine},
  author={Abeyrathna, K. Darshana and Granmo, Ole-Christoffer and Goodwin, Matthew},
  journal={arXiv preprint arXiv:2104.01409},
  year={2021}
}

@article{yadav2022fuzzy,
  title={The Fuzzy Tsetlin Machine},
  author={Yadav, Vikas and Granmo, Ole-Christoffer and Jiao, Lei},
  journal={arXiv preprint arXiv:2201.04977},
  year={2022}
}

@inproceedings{yazidi2022reltm,
  title={Relational Tsetlin Machines},
  author={Yazidi, Anis and Granmo, Ole-Christoffer and Jiao, Lei},
  booktitle={International Workshop on Mining and Learning with Graphs (MLG)},
  year={2022}
}

@inproceedings{tarasyuk2024mltm,
  title={Multi-Layer Tsetlin Machine: Architecture and Performance Evaluation},
  author={Tarasyuk, O. and Gorbenko, A. and Shafik, R. and Yakovlev, A.},
  booktitle={International Symposium on Tsetlin Machines (ISTM)},
  year={2024}
}

@article{Qi2023FedTMMA,
  title={FedTM: Memory and Communication Efficient Federated Learning with Tsetlin Machine},
  author={Shannon How Shi Qi and Jagmohan Chauhan and Geoff V. Merrett and Jonathan Hare},
  journal={2023 International Symposium on the Tsetlin Machine (ISTM)},
  year={2023},
  pages={1-8},
  url={https://api.semanticscholar.org/CorpusID:268254400}
}

@misc{gohari2024tpfltsetlinpersonalizedfederatedlearning,
      title={TPFL: Tsetlin-Personalized Federated Learning with Confidence-Based Clustering}, 
      author={Rasoul Jafari Gohari and Laya Aliahmadipour and Ezat Valipour},
      year={2024},
      eprint={2409.10392},
      archivePrefix={arXiv},
      primaryClass={cs.DC},
      url={https://arxiv.org/abs/2409.10392}, 
}

@inproceedings{qi2025fedtmos,
title={Fed{TMOS}: Efficient One-Shot Federated Learning with Tsetlin Machine},
author={Shannon How Shi Qi and Jagmohan Chauhan and Geoff V. Merrett and Jonathon Hare},
booktitle={The Thirteenth International Conference on Learning Representations},
year={2025},
url={https://openreview.net/forum?id=44hcrfzydU}
}

@ARTICLE{4787122,
  author={Aysal, Tuncer Can and Yildiz, Mehmet Ercan and Sarwate, Anand D. and Scaglione, Anna},
  journal={IEEE Transactions on Signal Processing}, 
  title={Broadcast Gossip Algorithms for Consensus}, 
  year={2009},
  volume={57},
  number={7},
  pages={2748-2761},
  doi={10.1109/TSP.2009.2016247}}

@article{Lei2020FromAT,
  title={From Arithmetic to Logic based AI: A Comparative Analysis of Neural Networks and Tsetlin Machine},
  author={Jie Lei and Adrian Wheeldon and Rishad A. Shafik and Alex Yakovlev and Ole-Christoffer Granmo},
  journal={2020 27th IEEE International Conference on Electronics, Circuits and Systems (ICECS)},
  year={2020},
  pages={1-4},
  url={https://api.semanticscholar.org/CorpusID:230513121}
}

@article{Zhao02112018,
author = {Liang Zhao and Wen-Zhan Song and Xiaojing Ye and Yujie Gu},
title = {Asynchronous broadcast-based decentralized learning in sensor networks},
journal = {International Journal of Parallel, Emergent and Distributed Systems},
volume = {33},
number = {6},
pages = {589--607},
year = {2018},
publisher = {Taylor \& Francis},
doi = {10.1080/17445760.2017.1294690},
URL = {https://doi.org/10.1080/17445760.2017.1294690},
eprint = {https://doi.org/10.1080/17445760.2017.1294690}
}

@book{2821576,
author = {Lynch, Nancy A.},
title = {Distributed Algorithms},
year = {1996},
isbn = {9780080504704},
publisher = {Morgan Kaufmann Publishers Inc.},
address = {San Francisco, CA, USA}
}

@book{355459,
author = {Peleg, David},
title = {Distributed computing: a locality-sensitive approach},
year = {2000},
isbn = {0898714648},
publisher = {Society for Industrial and Applied Mathematics},
address = {USA}
}

@INPROCEEDINGS{1238221,
  author={Kempe, D. and Dobra, A. and Gehrke, J.},
  booktitle={44th Annual IEEE Symposium on Foundations of Computer Science, 2003. Proceedings.}, 
  title={Gossip-based computation of aggregate information}, 
  year={2003},
  volume={},
  number={},
  pages={482-491},
  doi={10.1109/SFCS.2003.1238221}}

@article{lecun2010mnist,
         title={MNIST handwritten digit database},
         author={LeCun, Yann and Cortes, Corinna and Burges, CJ},
         journal={ATT Labs [Online]. Available: http://yann.lecun.com/exdb/mnist},
         volume={2},
         year={2010}
}

@article{Xiao2017FashionMNISTAN,
  title={Fashion-MNIST: a Novel Image Dataset for Benchmarking Machine Learning Algorithms},
  author={Han Xiao and Kashif Rasul and Roland Vollgraf},
  journal={ArXiv},
  year={2017},
  volume={abs/1708.07747},
  url={https://api.semanticscholar.org/CorpusID:702279}
}

@article{AbhishekV2022FederatedLC,
  title={Federated Learning: Collaborative Machine Learning without
Centralized Training Data},
  author={A AbhishekV and S Binny and R JohanT and Nithin Raj and Vishal Thomas},
  journal={international journal of engineering technology and management sciences},
  year={2022},
  url={https://api.semanticscholar.org/CorpusID:251659795}
}

@misc{akhtarshenas2024federatedlearningcuttingedgesurvey,
      title={Federated Learning: A Cutting-Edge Survey of the Latest Advancements and Applications}, 
      author={Azim Akhtarshenas and Mohammad Ali Vahedifar and Navid Ayoobi and Behrouz Maham and Tohid Alizadeh and Sina Ebrahimi and David López-Pérez},
      year={2024},
      eprint={2310.05269},
      archivePrefix={arXiv},
      primaryClass={cs.LG},
      url={https://arxiv.org/abs/2310.05269}, 
}

@misc{li2019fedmdheterogenousfederatedlearning,
      title={FedMD: Heterogenous Federated Learning via Model Distillation}, 
      author={Daliang Li and Junpu Wang},
      year={2019},
      eprint={1910.03581},
      archivePrefix={arXiv},
      primaryClass={cs.LG},
      url={https://arxiv.org/abs/1910.03581}, 
}

@misc{mcmahan2023communicationefficientlearningdeepnetworks,
      title={Communication-Efficient Learning of Deep Networks from Decentralized Data}, 
      author={H. Brendan McMahan and Eider Moore and Daniel Ramage and Seth Hampson and Blaise Agüera y Arcas},
      year={2023},
      eprint={1602.05629},
      archivePrefix={arXiv},
      primaryClass={cs.LG},
      url={https://arxiv.org/abs/1602.05629}, 
}

@misc{yang2019federatedmachinelearningconcept,
      title={Federated Machine Learning: Concept and Applications}, 
      author={Qiang Yang and Yang Liu and Tianjian Chen and Yongxin Tong},
      year={2019},
      eprint={1902.04885},
      archivePrefix={arXiv},
      primaryClass={cs.AI},
      url={https://arxiv.org/abs/1902.04885}, 
}

@misc{kairouz2021advancesopenproblemsfederated,
      title={Advances and Open Problems in Federated Learning}, 
      author={Peter Kairouz and H. Brendan McMahan and Brendan Avent and Aurélien Bellet and Mehdi Bennis and Arjun Nitin Bhagoji and Kallista Bonawitz and Zachary Charles and Graham Cormode and Rachel Cummings and Rafael G. L. D'Oliveira and Hubert Eichner and Salim El Rouayheb and David Evans and Josh Gardner and Zachary Garrett and Adrià Gascón and Badih Ghazi and Phillip B. Gibbons and Marco Gruteser and Zaid Harchaoui and Chaoyang He and Lie He and Zhouyuan Huo and Ben Hutchinson and Justin Hsu and Martin Jaggi and Tara Javidi and Gauri Joshi and Mikhail Khodak and Jakub Konečný and Aleksandra Korolova and Farinaz Koushanfar and Sanmi Koyejo and Tancrède Lepoint and Yang Liu and Prateek Mittal and Mehryar Mohri and Richard Nock and Ayfer Özgür and Rasmus Pagh and Mariana Raykova and Hang Qi and Daniel Ramage and Ramesh Raskar and Dawn Song and Weikang Song and Sebastian U. Stich and Ziteng Sun and Ananda Theertha Suresh and Florian Tramèr and Praneeth Vepakomma and Jianyu Wang and Li Xiong and Zheng Xu and Qiang Yang and Felix X. Yu and Han Yu and Sen Zhao},
      year={2021},
      eprint={1912.04977},
      archivePrefix={arXiv},
      primaryClass={cs.LG},
      url={https://arxiv.org/abs/1912.04977}, 
}

@misc{granmo2023tmcompositesplugandplaycollaborationspecialized,
      title={TMComposites: Plug-and-Play Collaboration Between Specialized Tsetlin Machines}, 
      author={Ole-Christoffer Granmo},
      year={2023},
      eprint={2309.04801},
      archivePrefix={arXiv},
      primaryClass={cs.CV},
      url={https://arxiv.org/abs/2309.04801}, 
}

@misc{wu2025verticalfederatedlearningpractice,
      title={Vertical Federated Learning in Practice: The Good, the Bad, and the Ugly}, 
      author={Zhaomin Wu and Zhen Qin and Junyi Hou and Haodong Zhao and Qinbin Li and Bingsheng He and Lixin Fan},
      year={2025},
      eprint={2502.08160},
      archivePrefix={arXiv},
      primaryClass={cs.LG},
      url={https://arxiv.org/abs/2502.08160}, 
}

@article{JMLR:v22:20-815,
  author  = {Yang Liu and Tao Fan and Tianjian Chen and Qian Xu and Qiang Yang},
  title   = {FATE: An Industrial Grade Platform for Collaborative Learning With Data Protection},
  journal = {Journal of Machine Learning Research},
  year    = {2021},
  volume  = {22},
  number  = {226},
  pages   = {1--6},
  url     = {http://jmlr.org/papers/v22/20-815.html}
}

@inproceedings{Li_2024,
   title={Multi-modal Anchor Gated Transformer with Knowledge Distillation for Emotion Recognition in Conversation},
   url={http://dx.doi.org/10.24963/ijcai.2024/905},
   DOI={10.24963/ijcai.2024/905},
   booktitle={Proceedings of the Thirty-ThirdInternational Joint Conference on Artificial Intelligence},
   publisher={International Joint Conferences on Artificial Intelligence Organization},
   author={Li, Jie and Ding, Shifei and Guo, Lili and Li, Xuan},
   year={2024},
   month=Aug, pages={8141–8149},
   collection={IJCAI-2024} }

@inproceedings{Huang_2023, 
   title={Vertical Federated Knowledge Transfer via Representation Distillation for Healthcare Collaboration Networks},
   url={http://dx.doi.org/10.1145/3543507.3583874},
   DOI={10.1145/3543507.3583874},
   booktitle={Proceedings of the ACM Web Conference 2023},
   publisher={ACM},
   author={Huang, Chung-ju and Wang, Leye and Han, Xiao},
   year={2023},
   month=Apr, pages={4188–4199},
   collection={WWW ’23} }

@INPROCEEDINGS{9923830,
  author={Rahman, Tousif and Wheeldon, Adrian and Shafik, Rishad and Yakovlev, Alex and Lei, Jie and Granmo, Ole-Christoffer and Das, Shidhartha},
  booktitle={2022 International Symposium on the Tsetlin Machine (ISTM)}, 
  title={Data Booleanization for Energy Efficient On-Chip Learning using Logic Driven AI}, 
  year={2022},
  volume={},
  number={},
  pages={29-36},
  doi={10.1109/ISTM54910.2022.00014}}

@misc{granmo2026tsetlinmachinegoesdeep,
      title={The Tsetlin Machine Goes Deep: Logical Learning and Reasoning With Graphs}, 
      author={Ole-Christoffer Granmo and Youmna Abdelwahab and Per-Arne Andersen and Karl Audun K. Borgersen and Paul F. A. Clarke and Kunal Dumbre and Ylva Grønningsæter and Vojtech Halenka and Runar Helin and Lei Jiao and Ahmed Khalid and Rebekka Omslandseter and Rupsa Saha and Mayur Shende and Xuan Zhang},
      year={2026},
      eprint={2507.14874},
      archivePrefix={arXiv},
      primaryClass={cs.LG},
      url={https://arxiv.org/abs/2507.14874}, 
}

@misc{pyTsetlinMachine,
  author = {Ole-Christoffer Granmo},
  title  = {pyTsetlinMachine},
  year   = {2019},
  url    = {https://github.com/cair/pyTsetlinMachine}
}

\end{document}